\documentclass[journal]{IEEEtran}
\usepackage{url}
\usepackage{xcolor}
\usepackage[mathlines]{lineno}
\usepackage[dvips]{epsfig}
\usepackage{lettrine}
\usepackage{float}
\usepackage{amsmath, amssymb}
\usepackage[tight,footnotesize]{subfigure}
\usepackage{multirow}
\usepackage{dsfont}
\usepackage{float}
\usepackage{algorithm}
\usepackage{algorithmic}
\usepackage{booktabs}
\usepackage{amsthm}
\usepackage{lineno}
\usepackage[colorlinks, linkcolor=black, urlcolor=blue,anchorcolor=black, citecolor=blue]{hyperref}

\newtheorem{definition}{\bf Definition}
\newtheorem{lemma}{\bf Lemma}
\newtheorem{assumption}{\bf Assumption}

\newcommand{\figurenames}{Figs.}

\ifCLASSINFOpdf
\else
\fi
\begin{document}
%
\title{A Unified Framework for Representation-Based Subspace Clustering of Out-of-Sample and Large-Scale Data}
%
%
%

\author{Xi Peng,
        Huajin Tang,~\IEEEmembership{Member~IEEE,}
        Lei Zhang,~\IEEEmembership{Member~IEEE,}，
        Zhang Yi,~\IEEEmembership{Senior Member~IEEE},
        Shijie Xiao~\IEEEmembership{Student Member~IEEE,} 
\thanks{X. Peng is are with Institute for Infocomm Research, A*STAR, Singapore 138632 (E-mail: pangsaai@gmail.com).}
\thanks{H. Tang, L. Zhang and Z. Yi are with the College of Computer
Science, Sichuan University, Chengdu, China 610065 (E-mail: htang@i2r.a-star.edu.sg, leizhang@scu.edu.cn, zhangyi@scu.edu.cn).}
\thanks{S.~J. Xiao is with School of Computer Engineering, Nanyang Technological University, Singapore (E-mail:xiao0050@ntu.edu.sg).}
\thanks{Corresponding author: H. Tang (E-mail: htang@i2r.a-star.edu.sg).}}

%
%

\markboth{Journal of \LaTeX\ Class Files,~Vol.~13, No.~9, September~2014}%
{Shell \MakeLowercase{\textit{et al.}}: Bare Demo of IEEEtran.cls for Journals}
%



\maketitle

\begin{abstract}
Under the framework of spectral clustering, the key of subspace clustering is building a similarity graph which describes the neighborhood relations among data points. Some recent works build the graph using sparse, low-rank, and $\ell_2$-norm-based representation, and have achieved state-of-the-art performance. However, these methods have suffered from the following two limitations. First, the time complexities of these methods are at least proportional to the cube of the data size, which make those methods inefficient for solving large-scale problems. Second, they cannot cope with out-of-sample data that are not used to construct the similarity graph. To cluster each out-of-sample datum, the methods have to recalculate the similarity graph and the cluster membership of the whole data set. In this paper, we propose a unified framework which makes representation-based subspace clustering algorithms feasible to cluster both out-of-sample and large-scale data. Under our framework, the large-scale problem is tackled by converting it as out-of-sample problem in the manner of ``sampling, clustering, coding, and classifying''. Furthermore, we give an estimation for the error bounds by treating each subspace as a point in a hyperspace. Extensive experimental results on various benchmark data sets show that our methods outperform several recently-proposed scalable methods in clustering large-scale data set.
\end{abstract}


%
\IEEEpeerreviewmaketitle

\section{Introduction}
\label{sec1}

\IEEEPARstart{C}{lustering} analysis aims to group  similar patterns into the same cluster by maximizing the inter-cluster dissimilarity and the intra-cluster similarity.  Over the past two decades, a number of clustering approaches have been proposed, for example, partitioning-based clustering~\cite{Yu2015Generalized}, kernel-based clustering~\cite{Muller2001}, and subspace clustering~\cite{Vidal2011}.

Subspace clustering aims at finding a low-dimensional subspace to fit each group of data points. It mainly contains two tasks, \emph{i.e.}, projecting the data set into another space (encoding) and calculating the cluster membership of the data set in the projection space (clustering). Popular subspace clustering methods include but not limit to statistical methods~\cite{Ma2007,Rao2008} and spectral clustering~\cite{Ng2002,Hou2014}. Spectral clustering finds the cluster membership of the data points by using the spectrum of an affinity matrix. The affinity matrix corresponds to a similarity graph of which each vertex denotes a data point, with the edge weights representing the similarities between connected points. Thus, at the heart of the spectral clustering is a similarity graph construction problem.

\begin{figure*}[t]
\centering
\includegraphics[width=0.96\textwidth]{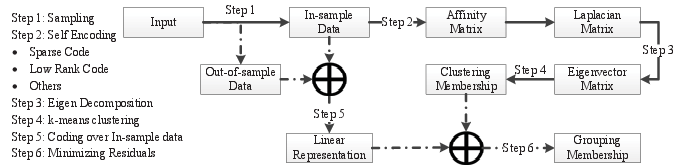}
\caption{\label{fig:1} Architecture of the proposed framework for scalable representation-based subspace clustering. The framework can be summarized as ``sampling (step 1), clustering (steps 2--4), coding (step 5), and classifying (step 6)''.
Solid and dotted lines are used to show the processes of clustering of in-sample data and out-of-sample data, respectively. For the out-of-sample problem, only steps 5 and 6 are needed.}
\end{figure*}

There are two widely-used approaches to build a similarity graph, \emph{i.e.}, Pairwise Distance (PD) and Reconstruction Coefficients (RC). Specifically, PD computes the similarity based on the distance (\emph{e.g.}, the Euclidean distance) between any two data points. However, PD cannot reflect the global structure of the data set, because its value only depends on connected data points. In contrast, RC denotes each data point as a linear combination of the other points and uses the representation coefficients as a similarity measurement. Several recent works have shown that RC is superior to PD in subspace clustering, for example, sparse representation~\cite{Elhamifar2009,Elhamifar2012,Cheng2010,Xu2012,Jing2013,Gao2013PAMI},
 low rank representation
~\cite{Liu2010,Liu2012,Favaro2011,Xiao2014},
  latent low rank representation~\cite{Liu2011}, and $\ell_2$-norm-based representation~\cite{Lu2012,peng2015robust}.

Although representation-based subspace clustering has  been extensively studied, how to solve the large-scale and out-of-sample clustering problems are less explored. Taking sparse subspace clustering (SSC)~\cite{Elhamifar2009,Elhamifar2012} as an example:  SSC iteratively computes the sparse codes of $n$ data points and performs eigen-decomposition over an $n\times n$ graph Laplacian matrix. Its computational complexity is more than $O(mn^3)$ even though the fastest $\ell_1$-solver is used, where $m$ denotes the dimensionality of the data set. Thus, any medium-sized data set will bring up large-scale problem with SSC. Moreover, SSC cannot handle out-of-sample data that are not used to construct the similarity graph. To cluster each previously unseen datum\footnote{In this paper, we assume that any previously unseen datum (\emph{i.e.}, out-of-sample datum) belongs to one of the subspaces spanned by in-sample data.}, SSC has to recompute the similarity graph and the cluster membership of the whole data set. In fact, most representation-based subspace clustering methods
~\cite{Liu2010,Liu2012,Xiao2014,Liu2011,Lu2012,Liu2012FRR} have suffered from similar limitations when dealing with large-scale or out-of-sample data.

To address such issues, we propose a unified framework for the representation-based subspace clustering algorithms. Our framework treats the large-scale problem as the out-of-sample problem in the manner of ``sampling, clustering, coding, and classifying'' (\figurename~\ref{fig:1}). Specifically, we split a large scale data set into two parts, in-sample data $\mathbf{X}$ and out-of-sample data $\mathbf{Y}$.~Then, we obtain the cluster membership of $\mathbf{X}$ and assign each out-of-sample datum to the nearest subspace spanned by $\mathbf{X}$. Under our framework, three scalable methods are presented, \emph{i.e.}, scalable sparse subspace clustering (SSSC), scalable low rank representation (SLRR), and scalable least square regression (SLSR). The proposed methods remarkably improve the computational efficiency of the original approaches while preserving a good clustering performance.

This paper is a substantial extension of our conference paper~\cite{Peng2013SSSC}, which is further improved from the following aspects: 1) We perform error analysis for our framework by treating each subspace in a well-defined hyperspace. The presented lower and upper error bounds are helpful in understanding the working mechanism of the nearest subspace classifier (specifically, sparse representation based classifier (SRC)~\cite{Wright2009}. To the best of our knowledge, this is the first work to perform errors analysis for SRC. 2) We additionally propose two scalable methods, \emph{i.e.}, SLRR and SLSR, which make low rank representation (LRR)~\cite{Liu2012} and least square regression (LSR)~\cite{Lu2012} feasible to cluster large scale data and out-of-sample data. 3) We perform extensive experiments to compare our methods with more scalable clustering methods on more data sets. 4)  We conduct comprehensive analysis for our approaches, including the performance with different out-of-sample grouping strategies and the influence of different parameters.

The rest of the paper is organized as follows: We provide in Section~\ref{sec2} a brief review of the representation-based clustering algorithms and some scalable spectral clustering methods. In Section~\ref{sec3}, we propose our framework and three scalable representation-based clustering algorithms, and further present some theoretical results on the error bound analysis of our framework. To demonstrate the performance of our proposed methods, we compare them with five recently-proposed scalable clustering approaches on nine data sets in Section~\ref{sec4}. Lastly, we give the conclusions and the further work in Section~\ref{sec5}.

\section{Representation-based Subspace Clustering}
\label{sec2}

In this paper, we use \textbf{lower-case bold letters} to represent column vectors and \textbf{UPPER-CASE BOLD LETTERS} to represent matrices. $\mathbf{A}^T$ and $\mathbf{A}^{-1}$ denote the transpose and pseudo-inverse of the matrix $\mathbf{A}$, respectively. $\mathbf{I}$ denotes the identity matrix.  Table~\ref{tab:1} summarizes some notations used throughout the paper.

\begin{table}[t]
\caption{Notations.}
\label{tab:1}
\begin{tabular}{ll}
\toprule
Notation	                              & Definition\\
\midrule
$n$ & the number of data points\\
$m$ & the dimensionality of a given data set\\
$k$ & the number of clusters\\
$p$ & the number of in-sample data\\
$t$ & the number of iterations of algorithm\\
$r$ & the rank of a given data matrix\\
$f(\mathbf{x}_{i})$ & the prediction for a given $\mathbf{x}_{i}$\\
$\mathbf{D}=[\mathbf{d}_1, \mathbf{d}_2, \ldots, \mathbf{d}_n]$ & data set\\
$[\mathbf{D}]_i$ & the data points belonging to the  subspace $S_i$\\
$\mathbf{X}=[\mathbf{x}_1, \mathbf{x}_2, \ldots, \mathbf{x}_p]$ & in-sample data\\
$\mathbf{Y}=[\mathbf{y}_1, \mathbf{y}_2, \ldots, \mathbf{y}_{n-p}]$ & out-of-sample data\\
$\mathbf{C}=[\mathbf{c}_1, \mathbf{c}_2, \ldots]$ & the representation of a given data set\\
$\mathbf{A}\in \mathds{R}^{n\times n}$ & affinity matrix based on $\mathbf{C}$\\
$\mathbf{L}\in \mathds{R}^{n\times n}$ & Laplacian matrix\\
$\mathbf{V}\in \mathds{R}^{n\times k}$ & eigenvector matrix\\
\bottomrule
\end{tabular}
\end{table}

\subsection{Sparse Representation Based Subspace Clustering}
\label{sec2.1}

Recently, Elhamifar and Vidal~\cite{Elhamifar2009,Elhamifar2012} proposed SSC  with well-founded recovery theory for independent subspaces and disjoint subspaces. SSC calculates the similarity among data points by solving the following optimization problem:
\begin{align}
\label{sec2equ:1}
&\underset{\mathbf{C},\mathbf{E},\mathbf{Z}}{\min} \hspace{1mm} \|\mathbf{C}\|_1+\lambda_{E}\|\mathbf{E}\|_1+\lambda_{Z}\|\mathbf{Z}\|_F \notag\\
&\mathrm{s.t.} \hspace{1mm} \mathbf{D}=\mathbf{D} \mathbf{C}+\mathbf{E}+\mathbf{Z},  \hspace{1mm} \mathrm{diag}(\mathbf{C})=0,
\end{align}
where $\mathbf{C}\in \mathds{R}^{n\times n}$ is the sparse representation of the data set $\mathbf{D}\in \mathds{R}^{m\times n}$, $\mathbf{E}$ corresponds to the sparse outlying entries, $\mathbf{Z}$ denotes the reconstruction errors for the limited representational capability, and the parameters $\lambda_{E}$ and $\lambda_{Z}$ balance these three terms in the objective function. (\ref{sec2equ:1}) is convex and can be solved by a number of $\ell_1$-solvers~\cite{Yang2010}.
After getting $\mathbf{C}$, SSC builds a similarity graph via $\mathbf{A}=\left|\mathbf{C}\right|^{T}+\left|\mathbf{C}\right|$ and performs spectral clustering~\cite{Ng2002} over the graph.

SSC is effective but inefficient. It needs $O(tn^2m^2 + tmn^3)$ to build the similarity graph even if the fastest $\ell_1$-solver is used, where $t$ denotes the number of iterations of the solver. In addition, SSC takes $O(n^3)$ to calculate the eigenvectors of the Laplacian matrix $\mathbf{L}$. Considering that $\mathbf{L}$ is a sparse matrix, the time complexity of this step could be reduced to $O(mn+mn^2)$ when Lanczos eigensolver is used. However, it is still a daunting task even for a moderate $n>100,000$.

\subsection{Low Rank Representation Based Subspace Clustering}
\label{sec2.2}

Different from SSC, LRR~\cite{Liu2012,Liu2011,Xiao_2015_CVPR} uses the  
lowest rank representation rather than the sparsest representation to build the similarity graph. The objective function of LRR is

\begin{equation}
\label{sec2equ:2}
\underset{\mathbf{C},\mathbf{E}}{\min}\hspace{1mm} \|\mathbf{C}\|_{\ast}+\lambda\|\mathbf{E}\|_{\ell} \hspace{3mm}
\mathrm{s.t.} \hspace{1mm} \mathbf{D}=\mathbf{DC}+\mathbf{E},
\end{equation}
where $\|\cdot\|_{\ast}$ denotes the nuclear norm, $\|\cdot\|_{\ell}$ could be chosen as $\ell_{2,1}$-norm, $\ell_{1}$-norm, or Frobenius norm,~depending on prior knowledge of the error structure. Generally, $\ell_{2,1}$-norm is adopted to deal with sample-specific corruption and outlier, $\ell_1$-norm is used to characterize the random corruption, and Frobenius norm is used to handle the Gaussian noise.

LRR, which adopts augmented Lagrange multipliers (ALM) method to solve (\ref{sec2equ:2}), takes $O(m^2 n + n^3)$ to perform singular value decomposition (SVD) over a dense matrix at each iteration. In addition, LRR will take $O(n^3+t_2 nk^2)$ to perform clustering, where $t_2$ denotes the number of iterations of the k-means method. Therefore, the overall time complexity of LRR is $O(t_{1}m^2 n+t_{1}n^3+t_2nk^2)$, where $t_1$ is the number of iterations of ALM.

\subsection{$\ell_2$-norm Based Methods}
\label{sec2.3}

SSC, LRR, and their extensions solve a convex optimization problem, of which the computational complexities are very high. Recently, least square regression (LSR)~\cite{Lu2012} has shown that $\ell_2$-norm-based representation can achieve the competitive result with faster speed. LSR aims at solving
\begin{equation}
\label{sec2equ:3}
\underset{\mathbf{C}}{\min} \hspace{1mm} \|\mathbf{D}-\mathbf{DC}\|_{F}^{2}+\lambda\|\mathbf{C}\|_{F}^{2} \hspace{3mm}
\mathrm{s.t.} \hspace{1mm} \mathrm{diag}(\mathbf{C})=\mathbf{0},
\end{equation}
where $\|\cdot\|_{F}$ denotes the Frobenius norm, the non-negative real number $\lambda$ is used to avoid overfitting, and the constraint guarantees that the $i$-th coefficient over $\mathbf{d}_{i}\in\mathbf{D}$ is zero.

\begin{algorithm}[t]
    \caption{Representation based Subspace Clustering.}
    \label{algorithm1}
    \begin{algorithmic}[1]
    \REQUIRE A set of data points $\mathbf{D}\in \mathds{R}^{m\times n}$ and the number of clusters $k$.
    \STATE %
Obtain the representation $\mathbf{C}^*$ by solving (\ref{sec2equ:1}), (\ref{sec2equ:2}) or (\ref{sec2equ:3}).%
    \STATE Get the affinity matrix via %
$\mathbf{A}=\left|\mathbf{C}^*\right|^T+\left|\mathbf{C}^*\right|$.
    \STATE Construct a Laplacian matrix $\mathbf{L}=\mathbf{I}-\mathbf{B}^{-1/2} \mathbf{A} \mathbf{B}^{-1/2}$ using $\mathbf{A}$, where $\mathbf{B}=\mathrm{diag}\{b_{i}\}$ with $b_{i}=\sum_{j=1}^{n}\mathbf{A}_{ij}$.
    \STATE Obtain the eigenvector matrix $\mathbf{U}\in \mathds{R}^{n\times k}$ which consists of the first $k$ normalized eigenvectors of $\mathbf{L}$ corresponding to its $k$ smallest eigenvalues.
    \STATE Get the segmentations of the data by performing k-means algorithm over the rows of $\mathbf{U}$.
    \ENSURE The cluster assignment of $\mathbf{D}$.
    \end{algorithmic}
\end{algorithm}

Lu et al.~\cite{Lu2012} provides two solutions to (\ref{sec2equ:3}) and the computational complexities of these solutions are $O(m^2n)$ at least. Thus, the overall computational complexity of LSR is about $O(m^2n+n^3+tnk^2)$, where $t$ denotes the number of iterations of the k-means method. Clearly, LSR has also suffered from the large-scale problem as SSC and LRR did.

Besides the large scale clustering problem, SSC, LRR, and LSR have suffered from the out-of-sample problem, \emph{i.e.}, they cannot cope with the data that are not used to construct the similarity graph. For each previously unseen datum, SSC, LRR, and LSR have to perform the algorithm over the whole data set once again. This makes them impossible to cluster incremental data. SSC, LRR, and LSR are summarized in Algorithm~\ref{algorithm1}.

There are some methods have been proposed to  reduce the time cost of minimizing lowest-rank matrix~\cite{Liu2012AS,Zhang2013}. However, the methods mainly focused on speeding up the encoding process without consideration of the clustering process.

\subsection{Scalable Spectral Clustering Algorithms}
\label{sec2.4}

Recently, some works have focused on solving the large-scale clustering problem of the traditional spectral clustering. One natural way is to reduce the time cost of eigen-decomposition over the Laplacian matrix. For example,~\cite{Fowlkes2004} adopted Nystr\"{o}m method to get the approximation of the eigenvectors of the whole similarity matrix.~\cite{Chen2011a} solved the generalized eigenvalue problem in a distributed computing platform.

Another way is reducing the data size by replacing the original data with a small number of samples.~\cite{Yan2009} presented a fast spectral clustering algorithm by selecting some representative points from the input and got the cluster assignment based on the chosen samples.~\cite{Chen2011} proposed landmark-based spectral clustering algorithm. The algorithm chooses $p$ representative points as the landmarks and constructs a Laplacian matrix via $\mathbf{L}=\mathbf{A}^T\mathbf{A}$, where the element of $\mathbf{A}\in \mathds{R}^{p\times n}$ is the pairwise distance between the input data and the landmarks.~\cite{Wang2011} selects the landmarks by performing selective sampling technique and running spectral clustering over the chosen samples based on pairwise distance.~\cite{Nie2011} proposed spectral embedded clustering (SEC) which groups out-of-sample data in a linear projection space. The main difference among the above works is the method to handle out-of-sample data. Different from the above sampling-based method, Belabbas and Wolfe~\cite{Belabbas2009} proposed a quantization based method with theoretical justification to select in-sample data in a deterministic way. By extending the quantization based method with self-organizing maps (SOMs), Tasdemir~\cite{Tasdemir2012} recently proposed a novel method by utilizing the quantization property of SOMs and neural gas to handle the large scale data set. Extensive experimental studies show that this method has achieved impressive performance compared with sampling-based methods on a range of data sets. Although numerous works have been conducted on speeding up the pairwise distance based clustering methods, very few researches have been done to enhance the scalability of the representation based approaches.

\section{Scalable Subspace Clustering and Error Analysis}
\label{sec3}

In this section, we present our framework which makes the representation based subspace clustering methods feasible to handle large scale data and out-of-sample data. Our method treats the large-scale problem as the out-of-sample problem by taking the strategy of ``sampling, clustering, coding, and classifying''. The first two steps choose a small number of data points as in-sample data and calculate the cluster membership of them. The third and fourth steps find a low-dimensional subspace to fit each group of out-of-sample data and assign the data to the subspace that has the minimal residual. Note that, to solve the out-of-sample problem, only the last two steps are needed.

\begin{figure*}[t]
\centering
\subfigure []{\label{fig:2a}\includegraphics[width=0.151\textwidth]{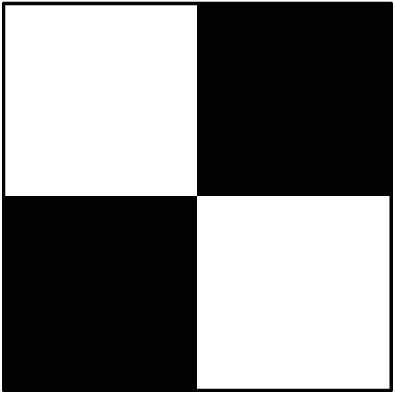}}\hspace{1mm}
\subfigure []{\label{fig:2b}\includegraphics[width=0.052\textwidth]{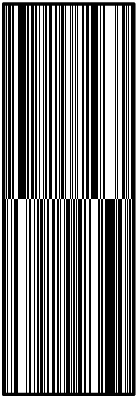}}\hspace{1mm}
\subfigure []{\label{fig:2c}\includegraphics[width=0.153\textwidth]{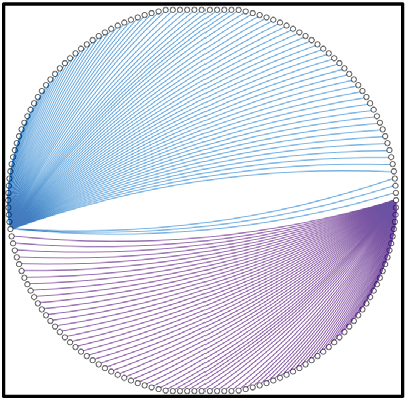}}\hspace{1mm}
\subfigure []{\label{fig:2d}\includegraphics[width=0.1515\textwidth]{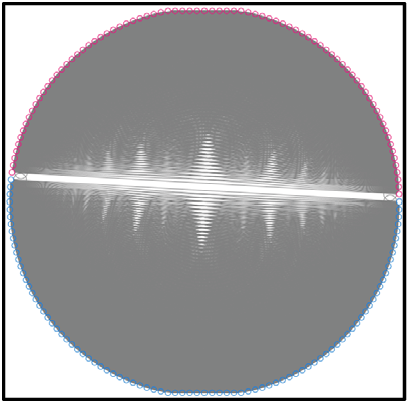}}\hspace{1mm}
\subfigure []{\label{fig:2e}\includegraphics[width=0.102\textwidth]{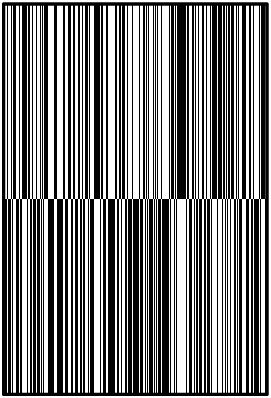}}\hspace{1mm}
\subfigure []{\label{fig:2f}\includegraphics[width=0.269\textwidth]{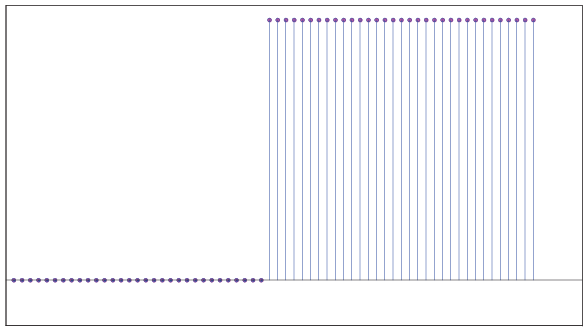}}
\caption{A toy example based on SSSC and SLRR for showing the effectiveness of our framework.
(a) A given data set $\mathbf{D}$ satisfying the sparsity assumption, where the rank of the data equals two;
(b) in-sample data $\mathbf{X}$ identifying using unique random sampling method. $\mathbf{X}$ and $\mathbf{D}$ are i.i.d.;
(c) The similarity graph of $\mathbf{X}$ achieved by SSSC;
(d) The similarity graph of $\mathbf{X}$ achieved by SLRR;
(e) out-of-sample data $\mathbf{Y}$ locating in the union of subspaces spanned by $\mathbf{X}$;
(f) The projection coefficients of an out-of-sample data point $\mathbf{y}\in S_{2}$, of which only the  coefficients over $S_{2}$ are nonzero. $\mathbf{y}$ is grouped into the subspace $S_{2}$ in terms of our method, which matches with the ground truth.
Under Assumption~\ref{asump:1}, this example shows that our framework can solve the large-scale and the out-of-sample problems for representation-based subspace clustering without loss of clustering quality.}
\label{fig:2}
\end{figure*}

\subsection{The Proposed Methods}
\label{sec3.1}

Our framework is based on a general assumption as follows:
\begin{assumption}
\label{asump:1}
Suppose the data set $[\mathbf{D}]_i \in \mathds{R}^{m\times n_i}$ is drawn from the subspace $S_i$,  one could use a small portion of $[\mathbf{D}]_i$, denoted by $[\mathbf{X}]_i\in \mathds{R}^{m\times p_i}$, to learn the structure of $S_i$, where $rank([\mathbf{D}]_i)=rank([\mathbf{X}]_i)$, $rank([\mathbf{X}]_i) \leq p_i \ll n_i$, and $S_i$ is a compact metric space.
\end{assumption}

Assumption~\ref{asump:1} is twofold. First, it implies that each data point could be encoded as a linear combination of a few basis (\emph{i.e.}, sparsity assumption). Second, it requires that $[\mathbf{X}]_i$ and $[\mathbf{D}]_i$ are independent and identically distributed (\emph{i.e.}, i.i.d.) so that out-of-sample data could be represented by $[\mathbf{X}]_i$. The assumption is very general on which most data mining and machine learning works are based.

In practice, the sparsity assumption is easily satisfied for high-dimensional data such as facial images. To satisfy the assumption of i.i.d., we need to find the representative points $\mathbf{X}\in \mathds{R}^{m\times p}$ from $\mathbf{D}\in \mathds{R}^{m\times n}$ so that out-of-sample data $\mathbf{Y}\in \mathds{R}^{m\times (n-p)}$ locate in the subspaces spanned by $\mathbf{X}$. To this end, some sampling techniques such as column selection method~\cite{Halko2011} can be used. However, these sampling methods are inefficient and cannot be applied to large scale setting. In this paper, we adopt uniform random sampling approach of which time cost is only $O(1)$. In addition to computational efficiency, the uniform random sampling method can perform comparably to the complex sampling techniques as shown in~\cite{Yan2009,Nie2011}. After sampling and getting the cluster membership of in-sample data $\mathbf{X}$, we handle out-of-sample data $\mathbf{Y}$ based on the knowledge learnt from $\mathbf{X}$. The simplest approach is assigning each $\mathbf{y}_{i}\in \mathbf{Y}$ to the nearest $\mathbf{x}_{j}\in \mathbf{X}$ in terms of the Euclidean distance. However, such approach implicitly requires some prior knowledge. For example, the data set must locate in the Euclidean space otherwise $\mathbf{y}_{i}$ would not be correctly clustered.

In this work, we compute the sparse representation of $\mathbf{Y}$ over $\mathbf{X}$ and assign each $\mathbf{y}_{i}$ to the nearest subspace based on SRC~\cite{Wright2009}. For each out-of-sample data point $\mathbf{y}_{i}$, the following optimization problem is solved
\begin{equation}
\label{equ:4} \min\hspace{1mm}\|\mathbf{c}_{i}\|_1
\hspace{3mm} \mathrm{s.t.} \hspace{1mm}
\|\mathbf{y}_{i}-\mathbf{X} \mathbf{c}_{i}\|_2<\epsilon,
\end{equation}
where $\epsilon>0$ is the error tolerance, $\mathbf{y}_{i}$ denotes an out-of-sample datum and $\mathbf{X}$ denotes in-sample data.

Once the optimal $\mathbf{c}_{i}$ is obtained, $\mathbf{y}_{i}$ is assigned to the nearest subspace which has the minimum residual by solving
\begin{equation}
\label{equ:5}
r_j(\mathbf{y}_{i})=\|\mathbf{y}_{i}-\mathbf{X}\mathbf{\delta}_{j}(\mathbf{c}_{i})\|_2.
\end{equation}
\begin{equation}
\label{equ:6}
f(\mathbf{y}_{i})=\mathop{\mathrm{argmin}}_{j}\{r_j(\mathbf{y}_{i})\},
\end{equation}
where the $f(\mathbf{y}_{i})$ denotes the assignment of $\mathbf{y}_{i}$, and the nonzero entries of $\mathbf{\delta}_{j}(\mathbf{c}_{i})\in\mathds{R}^p$ are the elements in $\mathbf{c}_{i}$ associating with the $j$-th subspace.

Although SRC has achieved a lot of successes in pattern recognition, some recent works~\cite{Zhang2011} showed that non-sparse representation can achieve comparable results with less time cost. Therefore, we perform linear coding scheme instead of sparse one by solving
\begin{equation}
\label{equ:7}
\hspace{6mm}
\mathop{\mathrm{min}}_{\mathbf{c}_{i}}\|\mathbf{y}_{i}-\mathbf{X}\mathbf{c}_{i}\|_2^2
+ \gamma \| \mathbf{c}_{i} \|_2^2,
\end{equation}
where $\gamma> 0$ is a positive real number. The second term is used to avoid over-fitting. Zhang et al.~\cite{Zhang2011} named this method as collaborative representation-based classification (CRC) and empirically showed that collaborative representation rather than the sparse one plays an important role in face recognition. After getting the coefficient of $\mathbf{y}_{i}$ via solving (\ref{equ:7}), $\mathbf{y}_{i}$ is assigned to the subspace that produces the minimal regularized residuals over all classes. Note that, (\ref{equ:7}) is also known as linear regression based classification \cite{Naseem2010} when $\gamma=0$.



Under our framework, SSSC, SLRR, and SLSR are proposed, which make SSC~\cite{Elhamifar2009,Elhamifar2012}, LRR~\cite{Liu2010,Liu2012}, and LSR~\cite{Lu2012} feasible to cluster large scale and out-of-sample data. Algorithm~\ref{algorithm3} summarizes our approaches and \figurename~\ref{fig:2} gives a toy example to show the effectiveness of our framework. In the example, we use the NodeXL software (a toolkit of Office)~\cite{Smith2010} to obtain the visualization of the similarity graphs  (see~\figurename~\ref{fig:2c} and \figurename~\ref{fig:2d}).

\begin{algorithm}[t]
    \caption{Scalable Sparse Subspace Clustering (SSSC), Scalable Low Rank Representation (SLRR), and Scalable Least Square Regression (SLSR).}
    \label{algorithm3}
    \begin{algorithmic}[1]
    \REQUIRE A given data set $\mathbf{D}\in \mathds{R}^{m\times n}$, the desired number of clusters $k$, and the rigid regression parameter $\gamma=10^{-6}$.
    \STATE Randomly select $p$ data points from $\mathbf{D}$ as in-sample data $\mathbf{X}=(\mathbf{x}_1,\mathbf{x}_2,\ldots,\mathbf{x}_p)$. The remaining samples are used as out-of-sample data $\mathbf{Y}=(\mathbf{y}_1,\mathbf{y}_2,\ldots,\mathbf{y}_{n-p})$.
    \STATE Perform SSC or LRR or LSR (Algorithm~\ref{algorithm1}) over $\mathbf{X}$ to get the cluster membership of $\mathbf{X}$.
    \STATE Project each out-of-sample data point $\mathbf{y}_{i}$ into the union of the subspaces spanned by $\mathbf{X}$ via solving
           \begin{equation}
           \label{equ:8}
             \mathbf{{c}}_{i}^\ast = (\mathbf{X}^T \mathbf{X}+\gamma\mathbf{I})^{-1}\mathbf{X}^T\mathbf{y}_{i}.
           \end{equation}
    \STATE Calculate the residuals of $\mathbf{y}_{i}$ over the $j$-th subspace by
           \begin{equation}
           \label{equ:9}
             r_j(\mathbf{y}_{i})=\|\mathbf{y}_{i}-\mathbf{X}\mathbf{\delta}_{j}(\mathbf{{c}}_{i}^\ast)\|_{2}.
           \end{equation}
    or the regularized residuals of $\mathbf{y}_{i}$ over all subspaces via solving
           \begin{equation}
           \label{equ:10}
             r_j(\mathbf{y}_{i})=\frac{\|\mathbf{y}_{i}-\mathbf{X}\mathbf{\delta}_{j}(\mathbf{{c}}_{i}^\ast)\|_{2}}{\|\mathbf{\delta}_{j}(\mathbf{{c}}_{i}^\ast)\|_{2}}.
           \end{equation}
    \STATE Assign $\mathbf{y}_{i}$ to the subspace which has the minimal residual by
           \begin{equation}
           \label{equ:11}
             f(\mathbf{y}_{i})=\mathop{\mathrm{argmin}}_{j}\{r_j(\mathbf{y}_{i})\}.
           \end{equation}
    \ENSURE The cluster membership of $\mathbf{D}$.
    \end{algorithmic}
\end{algorithm}

\subsection{Error Analysis}
\label{sec3.2}

In this section, we perform error analysis for the framework. Lemma~\ref{lem:1} shows that the clustering partitions solely based on in-sample data $\mathbf{X}\in\mathds{R}^{m\times p}$ will converge to the partitions based on the whole data set $\mathbf{D}\in\mathds{R}^{m\times n}$, when $n \to \infty$ and the sampled data is enough. Based on Lemma~\ref{lem:1}, we show that the error bound of our framework only depend on the grouping errors of out-of-sample data $\mathbf{Y}\in \mathds{R}^{m\times (n-p)}$. Moreover, Lemma~\ref{lem:2} is the preliminary step to our result.

\begin{lemma}[\cite{Luxburg2004}]
\label{lem:1}
Under Assumption~\ref{asump:1}, if the first $k$ eigenvalues of $\mathbf{L}_{\mathbf{D}}$ have multiplicity 1, then the same holds for the first $k$ eigenvalues of $\mathbf{L}_{\mathbf{X}}$ for sufficiently large $p$, where $\mathbf{L}_{\mathbf{D}}$ and $\mathbf{L}_{\mathbf{X}}$ denote the Laplacian matrix based on $\mathbf{D}$ and $\mathbf{X}$, respectively. In this case, the first $k$ eigenvalues of $\mathbf{L}_{\mathbf{X}}$ converge to the first $k$ eigenvalues of $\mathbf{L}_{\mathbf{D}}$, and the corresponding eigenvectors converge almost surely. The clustering partitions constructed by normalized spectral clustering from the first $k$ eigenvectors on finite samples converge almost surely to a limit partition of the whole data space.
\end{lemma}

From Lemma~\ref{lem:1}, we can find that the additive clustering error induced by our framework comes from the process of grouping out-of-sample data $\mathbf{Y}$. Thus, the problem becomes finding the error boundary of the Nearest Subspace (NS) classifier ((\ref{equ:9}) or (\ref{equ:10})).

The representation-based NS classifiers have been extensively studied in~\cite{Wright2009,Zhang2011,Gao2013TIP}, however, theoretical analysis on it receives little attention.~\cite{Wang2013} presents a theoretical explanation to SRC~\cite{Wright2009} from the view of maximizing performance margin. However, the error boundary of SRC is still unknown. In this paper, we mainly investigate the performance of SRC (\emph{i.e.}, (\ref{equ:9})) from theoretical perspective. To the best of our knowledge, this is the first work to analyze the error bounds for the NS classifiers.

It is challenging to perform error analysis on the NS classifiers because the active sets (the nonzero set of $\mathbf{c}$) of different data points are different. In other words, it is difficult to find an invariant set of support vectors to represent each subspace. Therefore, the classic margin analysis theory cannot be directly used to the NS classifiers. To solve this problem, we propose treating each subspace as a point in a hyperspace. We have the following definition.

\begin{definition}
\label{pro1}
The hyperspace $\mathcal{H}=\{S, \mathbf{y}\}$ is a set of subspaces, in which each subspace $S_{j}$ corresponds to a point and the distance between $\mathbf{y}_{i}$ and $S_{j}$ is defined as the residual $r_j(\mathbf{y}_{i})$.
\end{definition}

\begin{figure*}[t]
\centering
\subfigure []{\label{fig:3a}\includegraphics[width=0.3\textwidth]{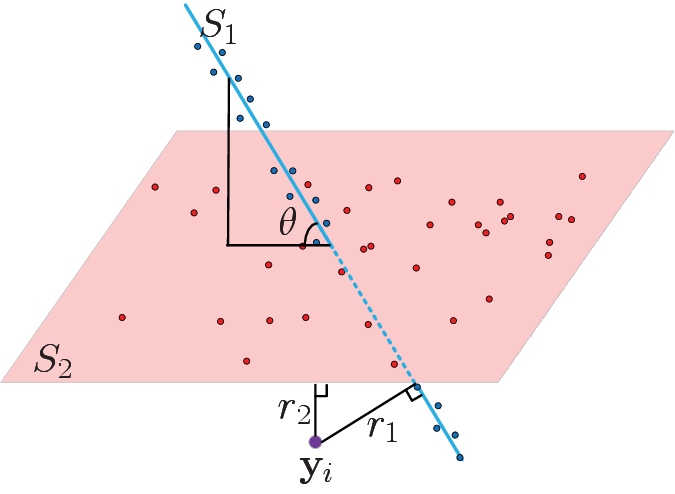}}\hspace{1cm}
\subfigure []{\label{fig:3b}\includegraphics[width=0.25\textwidth]{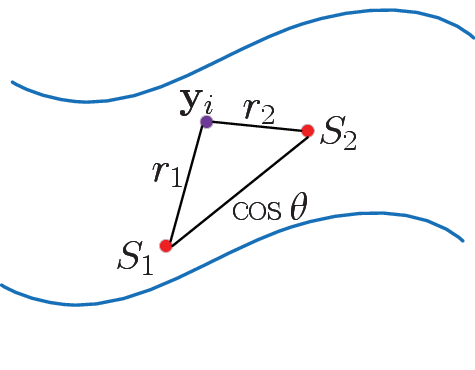}}\hspace{1cm}
\subfigure []{\label{fig:3c}\includegraphics[width=0.25\textwidth]{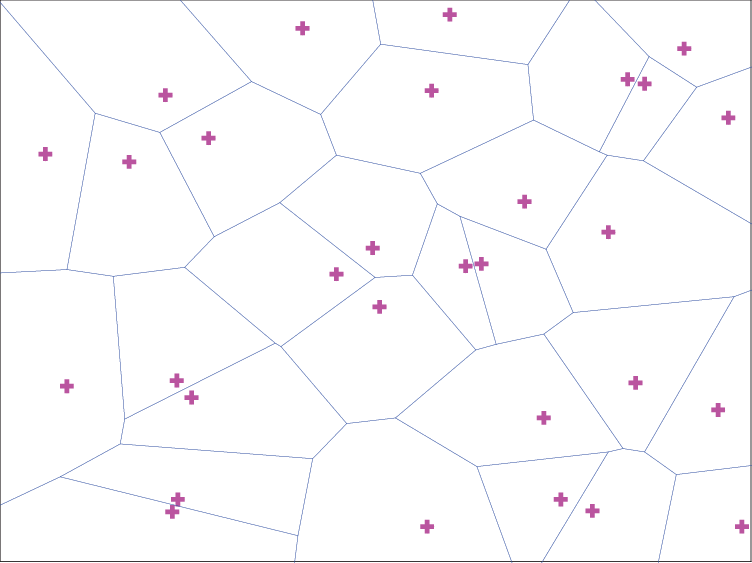}}
\caption{ (a) Two subspaces $S_1$ and $S_2$ spanned by in-sample data. We denote an out-of-sample data point by $\mathbf{y}_{i}$. $\theta$ is the principal angle between $S_1$ and $S_2$, and $r_1$ and $r_2$ are the residuals associating with $S_1$ and $S_2$. (b) The hyperspace in which $S_1$ and $S_2$ are regarded as two data points. (c) the decision boundary of the nearest subspace classifier in the hyperspace.}
\label{fig:3}
\end{figure*}

Based on the above definition, the NS classifier could be regarded as the nearest neighbor classifier in the hyperspace (see \figurename~\ref{fig:3}) so that one can avoid to find the support vectors for each category. Note that,~\cite{Hamm2008} treats each subspace as a data point in the Grassmann space in which the distance is defined as the principle angle between the subspaces. Clearly, the adopted distance metric is the major difference between Grassmann space and the above defined hyperspace. Indeed, Grassmann space can be regarded as a special case of the hyperspace, which will be further discussed at the end of this section.

\begin{lemma}[Cover-Hart inequality~\cite{Fukunaga1990}]
\label{lem:2}
For any distribution of $(\mathbf{Y}, g(\mathbf{Y}))$,
the asymptotic error $R$ of the nearest neighbor classifier is bounded by
\begin{equation}
\label{equ:12}
R^{\ast} \leq R \leq R^{\ast}\left(2-\frac{k}{k-1}R^{\ast}\right),
\end{equation}
where $g(\mathbf{Y})$ is the ground truth for $\mathbf{Y}$, $k$ denotes the number of subject, and $R^{\ast}$ denotes the Bayes error which is the lowest possible error rate for a given class of classifier.
\end{lemma}

Based on Lemma~\ref{lem:2}, the problem is equivalent to estimating the Bayes error in the defined hyperspace. Without loss of generality, we deal with the case of binary classification, \emph{i.e.}, $k=2$ and $f(\mathbf{y})=\{-1,1\}$.

\begin{lemma}
	 The error bound of the nearest subspace classifier
\begin{equation}
\label{equ:13}
 f(\mathbf{y}_{i})=\mathop{\mathrm{argmin}}_{j}\{\|\mathbf{y}_{i}-\mathbf{X}\mathbf{\delta}_{j} (\mathbf{c}_{i}^{\ast})\|\},
\end{equation}
is given by
\begin{equation}
\label{equ:14}
\frac{|1-\max(\alpha_{-1},\alpha_{1})|}{2+\alpha_{-1}+\alpha_{1}}\leq R
\leq \min(0.5,\frac{2+2\min(\alpha_{-1},\alpha_{1})}{|1-\alpha_{-1}|+|1-\alpha_{1}|)}),
\end{equation}
where $\mathbf{y}_{i}\in \mathds{R}^{m}$ is the input, $\mathbf{{c}}_{i}^\ast = \mathbf{\Theta}\mathbf{y}_{i}$, $\alpha_{j}=\|[\mathbf{X}]_{j} \mathbf{\Theta}\|_{F}$, $[\mathbf{X}]_{j}\in \mathbf{R}^{m \times p}$ replaces the elements of $\mathbf{X}$ with zeros unless the elements belong to $S_{j}$, $j=\{1,2\}$ denotes the index of subject, $\mathbf{\Theta}=(\mathbf{X}^T \mathbf{X}+\gamma\mathbf{I})^{-1}\mathbf{X}^T$, and the nonzero entries of $\mathbf{\delta}_{j}(\mathbf{c}_{i}^{\ast})\in\mathds{R}^p$ are the elements in $\mathbf{c}_{i}^{\ast}$ associated with the subspace $S_j$.
\end{lemma}

\begin{proof}

Let $\eta(\mathbf{y}_{i})$ be the conditional probability that the prediction for a given $\mathbf{y}_{i}$ is $1$, \emph{i.e.}, $\eta(\mathbf{y}_{i}) \triangleq p(f(\mathbf{y}_{i})=1|\mathbf{y}_i)$. In this case, the Bayes error $R^{\ast}$ for $\mathbf{y}_{i}$ is given by
\begin{align}
\label{equ:15}
R^{\ast}(\mathbf{y}_{i})&=\min \{\eta(\mathbf{y}_{i}), 1-\eta(\mathbf{y}_{i}) \}
\end{align}
According to (\ref{equ:15}), it is obvious that $0 \leq R^{\ast}(\mathbf{y}_{i}) \leq 0.5$.

We define the probability that $\mathbf{y}_{i}$ belongs to the subspace $S_{j}$ using the residual $r_{j}(\mathbf{y}_{i})=\|\mathbf{y}_{i}-\mathbf{X}\mathbf{\delta}_{j}(\mathbf{{c}}_{i}^\ast)\|_{2}$, \emph{i.e.},
\begin{equation}
\label{equ:16}
\eta(\mathbf{y}_{i})=1-\frac{r_{1}(\mathbf{y}_{i})}{\sum{r_{j}(\mathbf{y}_{i})}}.
\end{equation}

Let $\mathbf{\delta}_{j}(\mathbf{{c}}_{i}^\ast)=\mathbf{\Delta}_{j}\mathbf{c}_{i}^{\ast}$, where $\mathbf{\Delta}_{j}\in \mathds{R}^{p\times p}$ is a diagonal matrix of which nonzero diagonal entries indicate the columns of $\mathbf{X}$ belonging to the subspace $S_{j}$. Since $\mathbf{c}_{i}^{\ast}=\mathbf{\Theta}\mathbf{y}_{i}$, we have
\begin{equation}
\label{equ:17}
r_{j}(\mathbf{y}_{i})=\big\|  \mathbf{y}_{i} -  [\mathbf{X}]_{j} \mathbf{\Theta}  \mathbf{y}_{i}\big\|_{2},
\end{equation}
where $[\mathbf{X}]_{j}=\mathbf{\mathbf{X}}\mathbf{\Delta}_{j}$.

Thus, to find the bound of (\ref{equ:13}), we only need to identify the lower and upper bounds of $r_{j}(\mathbf{y}_{i})$.

Step 1: From the reverse triangle inequality of vector norm, we have
\begin{equation}
\label{equ:18}
r_{j}(\mathbf{y}_{i}) \geq  \big| \|\mathbf{y}_{i}\|_{2} - \|[\mathbf{X}]_{j} \mathbf{\Theta} \mathbf{y}_{i} \|_{2} \big|.
\end{equation}

For any vectors $\mathbf{x}$ and $\mathbf{y}$, Cauchy-Schwarz inequality suggests that $\|\mathbf{x}^{T}\mathbf{y}\|_{2} \leq \|\mathbf{x}\|_2 \|\mathbf{y}\|_2$. Since the Frobenius norm is subordinate to $\ell_2$-norm, (\ref{equ:18}) gives that
\begin{align}
\label{equ:19}
r_{j}(\mathbf{y}_{i}) & \geq \big|   \|\mathbf{y}_{i}\|_{2} - \|[\mathbf{X}]_{j} \mathbf{\Theta} \|_{F} \|\mathbf{y}_{i}\|_2 \big| \notag\\
& =\big| 1 -  \|[\mathbf{X}]_{j} \mathbf{\Theta} \|_{F}\big|  \|\mathbf{y}_{i}\|_{2},
\end{align}
where $\|\cdot\|_{F}$ denotes the Frobenius norm.

Step 2: For any vectors $\mathbf{x}$ and $\mathbf{y}$, it must hold that $\|\mathbf{x} - \mathbf{y}\|_{2} \leq \|\mathbf{x} + \mathbf{y}\|_{2} \leq \|\mathbf{x}\|_{2} + \|\mathbf{y}\|_{2}$. Thus, we have
\begin{align}
\label{equ:20}
r_{j}(\mathbf{y}_{i}) & = \big\| \mathbf{y}_{i} - [\mathbf{X}]_{j} \mathbf{\Theta} \mathbf{y}_{i} \big\|_{2} \notag\\
                &\leq \left\| \mathbf{y}_{i} \right\|_{2} + \left\| [\mathbf{X}]_{j}\mathbf{\Theta}\mathbf{y}_{i}    \right\|_{2} \notag\\
               &\leq \left\| \mathbf{y}_{i} \right\|_{2} + \left\| [\mathbf{X}]_{j}\mathbf{\Theta} \right\|_{F} \| \mathbf{y}_{i}\|_{2} \notag\\
               & = \big(1+\left\| [\mathbf{X}]_{j}\mathbf{\Theta} \right\|_{F} \big) \| \mathbf{y}_{i}\|_{2}
\end{align}

Let $\alpha_{j}=\big\| [\mathbf{X}]_{j} \mathbf{\Theta} \big\|_{F}$\footnote{In practice, we often normalize $\alpha_{j}$ via $\alpha_{j}=\alpha_{j}/\sum_{j}\alpha_{j}$.} and combine (\ref{equ:16}), (\ref{equ:19}), and (\ref{equ:20}), we have
\begin{equation}
\label{equ:21}
\frac{|1-\alpha_{-1}|}{2+\alpha_{-1}+\alpha_{1}} \leq\eta(\mathbf{y}_{i}) \leq \frac{1+\alpha_{-1}}{|1-\alpha_{-1}| + |1-\alpha_{1}|},
\end{equation}
and
\begin{equation}
\label{equ:22}
\frac{|1-\alpha_{1}|}{2+\alpha_{-1}+\alpha_{1}} \leq 1-\eta(\mathbf{y}_{i}) \leq \frac{1+\alpha_{1}}{|1-\alpha_{-1}|+|1-\alpha_{1}|},
\end{equation}
respectively.


Clearly, the boundary of the expected Bayes error $R^{\ast}=E\{ R^{\ast}(\mathbf{Y})\}$ is independent of the out-of-sample data $\mathbf{Y}$. From Lemma~\ref{lem:2}, the following relations hold:
\begin{equation}
\label{equ:22b}
R^{\ast} \leq R \leq 2R^{\ast}\left(1-R^{\ast}\right) \leq 2R^{\ast}.
\end{equation}

Since $0\leq R^{\ast}\leq 0.5$, then from (\ref{equ:21}), (\ref{equ:22}), and (\ref{equ:22b}), we have
\begin{align}
\label{equ:23}
\frac{|1-\max(\alpha_{-1},\alpha_{1})|}{2+\alpha_{-1}+\alpha_{1}}\leq R
\leq \min(0.5,\frac{2+2\min(\alpha_{-1},\alpha_{1})}{|1-\alpha_{-1}|+|1-\alpha_{1}|}).
\end{align}

This completes the proof.
\end{proof}

\begin{figure*}[t]
\centering
\subfigure [$\gamma=10^{-6}$]{\label{fig:4a}\includegraphics[width=0.42\textwidth]{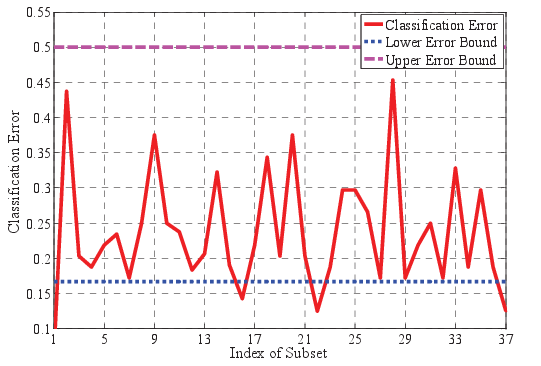}}\hspace{6mm}
\subfigure [$\gamma=10^{-6}$]{\label{fig:4b}\includegraphics[width=0.405\textwidth]{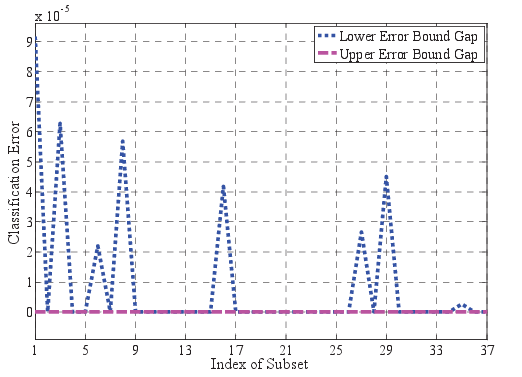}}
\subfigure [$\gamma=10^{-12}$]{\label{fig:4c}\includegraphics[width=0.42\textwidth]{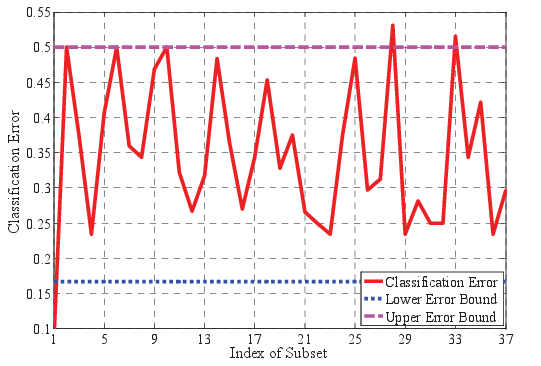}}\hspace{6mm}
\subfigure [$\gamma=10^{-12}$]{\label{fig:4d}\includegraphics[width=0.415\textwidth]{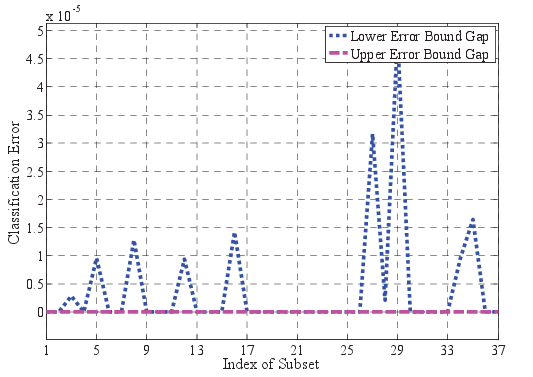}}
	\caption{A real-world example to validate the estimated error bounds. (a) and (c): Classification error and error bound of eq.(\ref{equ:13}) on 37 subsets of Extended Yale database B. Each subset consists of the samples from the first category and another category. (b) and (d): The gap between two different error bounds derived from equations (\ref{equ:22}) and (\ref{equ:25}). }
\label{fig:4}
\end{figure*}

From the above analysis, we can conclude that:
\begin{itemize}
  \item The error bound only depends on the structure of the subspaces spanned by in-sample data under Assumption~\ref{asump:1}. Indeed, the structure of the subspaces is also the unique factor to affect the clustering quality as shown in~\cite{Elhamifar2012, Liu2012}. Thus, we argue that our framework solves the large-scale and the out-of-sample problems for the representation-based subspace clustering methods without introducing new error factors. This is largely different from the traditional methods~\cite{Yan2009,Chitta2011} whose performance depends on the sampling rate.
   \item Considering $\mathbf{X}^{T}\mathbf{X}$ is well conditioned, then one sets $\gamma=0$. $\alpha_{j}=[\mathbf{X}]_{j}\mathbf{X}$ measures the similarity between the subspace $[\mathbf{X}]_{j}$ and $\mathbf{X}$ using their inner product. More generally (\emph{i.e.}, $\gamma> 0$), let $\theta_{i}$ be the $i$-th principal angle between $[\mathbf{X}]_{j}$ and $\mathbf{\Theta}$, then, it holds that $\sigma_{i}=\cos\theta_{i}$, where $\sigma_{i}$ is the $i$-th singular value of $[\mathbf{X}]_{j}\mathbf{\Theta}$. According to the definition of the Frobenius norm, \emph{i.e.}, $\alpha_{j}=\sqrt{\mathop{\sum}_{i}^{r_{j}}\sigma_{i}^{2} }$, we have $\alpha_{j}=\|[\mathbf{X}_{j}]\mathbf{\Theta}\|_{F}=\sqrt{\mathop{\sum}_{i}^{r_{j}}\cos^{2}\theta_{i} }$ which measures the distance between $[\mathbf{X}]_{j}$ and $\mathbf{\Theta}$ by their principal angles, where $r_{j}$ is the rank of $[\mathbf{X}]_{j}\mathbf{\Theta}$.
\end{itemize}

Under Assumption~\ref{asump:1}, our error analysis method is validate only when the following two conditions are satisfied when: 1) the data are sampled from two subspaces, \emph{i.e.}, $k=2$. If $k>2$, one may extend our method by recursively transforming the multiple clusters problem into binary one even though this task may need massive effort; and 2) in-sample data $[\mathbf{X}]_{j}$ have been correctly clustered. Otherwise, one needs identify the error bound for the whole framework not just for grouping out-of-sample data. The difficulty of this task is how to identify the influence of perturbation due to sampling. A possible way to solve this problem is perturbation theory that has been studied in quantum mechanics. However, this is beyond the main scope of this paper.

To validate our theoretical results, we perform experiments on 37 subsets of Extended Yale database B~\cite{Georghiades2001}. Each subset consists of the samples from the first category and one of the others. We use 64 (32 samples per subject) samples for training and the remaining samples for testing. Moreover, we use principle components analysis (PCA) as the preprocess step to extract 60 features from training and testing data.~\figurename~\ref{fig:4} shows results from which one can find that:
\begin{itemize}
  \item We successfully estimate the error bounds for 33 and 34 out of 37 subsets in the case of $\gamma=10^{-6}$ and $\gamma=10^{-12}$, respectively. The failure cases may be attributed to the following reasons: First, the classification error (solid line) is calculated based on training data and testing data, whereas the error bounds (dotted lines) are estimated only based on training data. When training data cannot represent the distribution of the whole data space, the estimated bounds will be incorrect. Second, our analysis is based on Assumption~\ref{asump:1}, which may not be perfectly satisfied by real-world data (\emph{e.g.},~the Extended Yale database B) since real-world data is often complex.
   \item \figurenames~\ref{fig:4a} and~\ref{fig:4c} show that a larger $\gamma$ may reduce the classification error rate, while increasing the failure rate of our error analysis method. The reason is that $\gamma$ is used to avoid overfitting by adding a value to the diagonal entries of $\mathbf{X}^{T}\mathbf{X}$, which actually affects the structure of the observed data.
   \item Besides (\ref{equ:22}), we derive another bound for $1-\eta(\mathbf{y}_{i})$ from (\ref{equ:21}) instead of (\ref{equ:16}), (\ref{equ:19}), and (\ref{equ:20}), \emph{i.e.},
			\begin{equation}
			\label{equ:25}
			1-\frac{1+\alpha_{-1}}{|1-\alpha_{-1}| + |1-\alpha_{1}|}  \leq 1-\eta(\mathbf{y}_{i}) \leq 1-  \frac{|1-\alpha_{-1}|}{2+\alpha_{-1}+\alpha_{1}}.
			\end{equation}
			\figurenames~\ref{fig:4b} and \ref{fig:4d} shows the gap between these two different formulations. Clearly, the gaps are close to zero.
\end{itemize}

\subsection{Complexity Analysis}
\label{sec3.4}

Suppose $p$ samples are selected from $n$ data points with dimensionality of $m$, SSSC needs $O(t_1 p^2m^2+t_1 mp^3+p^2+t_2pk^2)$ to get the cluster membership of in-sample data when the Homotopy optimizer~\cite{Osborne2000} is used to solve the $\ell_1$-minimization problem and the Lanczos eigensolver is used to compute the eigenvectors of $\mathbf{L}\in \mathds{R}^{p\times p}$, where $k$ is the number of clusters, and $t_1$ and $t_2$ is the number of iterations of Homotopy optimizer and the k-means algorithm, respectively. To group out-of-sample data points, SSSC needs to compute the pseudo-inverse of the an $m\times m$ matrix and calculate  the linear representation of $\mathbf{Y}\in \mathds{R}^{m\times (n-p)}$ in $O(pm^2+p^3+(n-p)p^2)$.

Putting everything together, the computational complexity of SSSC is $O(t_1mp^3+t_2pk^2+np^2)$ since $k, m <p\ll n$. Clearly, the cost of SSSC is largely less than that of SSC ($O(t_1mn^3+t_2nk^2)$). In the similar way, one can get the computational complexities of SLRR and SLSR. Table~\ref{tab:2} reports the computational complexities of our methods and the  original algorithms.

\begin{table}[t]
\caption{Computational complexity of SSC, LRR, LSR, and their scalable versions proposed in this paper. $t_1$, $t_2$, and $t_3$ correspond to the number of iterations of the $\ell_1$-solver, the rank-minimizer, and the k-means clustering method, respectively.}
\label{tab:2}
\begin{center}
\begin{tabular}{llc}
\toprule
Algorithms & Time Complexity & Space Complexity\\
\midrule
SSC~\cite{Elhamifar2009,Elhamifar2012} & $t_1mn^3+t_2nk^2$ & $mn^2$\\
SSSC & $t_1mp^3+t_2pk^2+np^2$ & $mp^2$ \\
LRR~\cite{Liu2010,Liu2012} & $t_3(m^2n+n^3)+t_2nk^2$ & $mn^2$ \\
SLRR & $t_3(pm^2+p^3)+t_2p^3+np^2$ & $mp^2$\\
LSR~\cite{Lu2012} & $m^2n+n^3+t_{2}nk$ & $mn^2$ \\
SLSR & $pm^2+np^2+t_{2}pk$ & $mp^2$ \\
\bottomrule
\end{tabular}
\end{center}
\end{table}

\section{Experimental Results}
\label{sec4}
In this section, we carry out some experiments to show the effectiveness and efficiency of SSSC, SLRR, and SLSR.

The experiments consist of five parts, Section~\ref{sec4.3} investigates the performance of our methods to the varying parameters; Section~\ref{sec4.4} reports the results of all the evaluated algorithms with different sampling rates; Section~\ref{sec4.5} compares our methods with the corresponding original algorithms on three facial data sets. Moreover, we also investigate the performance of two nearest subspace classifiers (\ref{equ:9}) and (\ref{equ:10}); Section~\ref{sec4.6} reports the clustering quality of the tested methods on three medium-sized data sets including facial images, handwritten digital data, and documental corpus; Section~\ref{sec4.7} shows the results on three large scale data sets.

\subsection{Data Sets}
\label{sec4.1}

We perform experiments on nine real-world data sets including facial images, handwritten digital data, news corpus, etc. The data sets consist of three small-sized data sets, three medium-sized data sets, and three large scale data sets. We presented some statistics of the data sets in Table~\ref{tab:3} and a brief description as follows.

In general, facial images are assumed to be located in the low-dimensional manifold. In the experiments, we investigate four popular facial data sets, \emph{i.e.}, AR~\cite{Martinez1998}, Extended Yale database B (ExYaleB)~\cite{Georghiades2001}, Labeled Faces in the Wild-a (LFW)~\cite{Gary2007}, and Multi-PIE (MPIE)~\cite{Gross2010}. AR includes over 4,000 face images of 126 people (70 male and 56 female).  In our implementation, we used a subset of AR which contains 1,400 clean faces randomly selected from 50 male subjects and 50 female subjects.
LFW contains 13,123 images captured from uncontrolled environment with variations of pose, illumination, expression, misalignment, and occlusion. We use a subset of the aligned LFW which includes 143 subjects with no less than 11 samples per subject. MPIE contains the facial images of 286 individuals captured in four sessions with simultaneous variations in pose, expression and illumination\footnote{illuminations of the used MPIE: {0,1,3,4,6,7,8,11,13,14,16,17,18,19}}. We use all frontal images from all the sessions. For computational efficiency, we downsize AR images from $165\times 120$ to $55\times 40$ ($1/9$), ExYaleB images from $192\times 168$ to $48\times 42$ ($1/16$), and MPIE images from $100\times 82$ to $50\times 41$ ($1/4$). Moreover, we perform PCA over the downsized data to retain 98\% energy. For each LFW image, ``divide and conquer'' strategy is adopted as did in~\cite{Yang2012-Relaxed}. In details, each image is partitioned into $2\times 2$ blocks; and then the discrimination-enhanced feature in each block is extracted; after that, all blocks' features are concatenated to form the final feature vector.

Reuters-21578 (RCV)~\cite{Cai2005} is a documental corpus. In the experiments, the first 785 principle components of RCV are extracted as the features. We also use three UCI data sets\footnote{\url{http://archive.ics.uci.edu/ml/datasets.html}}, \emph{i.e.}, PenDigits, Covtype~\cite{Alimoglu1996}, and PokerHand~\cite{Blackard1999}. PokerHand is an unbalanced data set, of which the maximal class contains 501,209 samples, compared with 3 samples of the minimal class. We examine the performance of the tested algorithms using the original data set (PokerHand-2) and a subset (PokerHand-1) with 971,329 data points from three largest subjects.

\begin{table}[t]
\caption{Data sets used in the experiments. The number in the parentheses denotes the retaining energy by PCA.}
\label{tab:3}
\begin{center}
\begin{tabular}{lrrlr}
\toprule
Data sets	                    &  \# samples                         & Dim.      & \# features    & \# classes  \\
\midrule
AR~\cite{Martinez1998}          &   1,400                                        & 19,800      & 167 (98\%)        & 100  \\
ExYaleB~\cite{Georghiades2001}  &   2,414                                        & 32,256      & 114 (98\%)        & 38   \\
LFW~\cite{Taigman2009} & 4,174 & 62,500 & 560 & 143\\
MPIE~\cite{Gross2010}           &   8,916                                        & 8,200       & 115 (98\%)        & 286  \\
RCV~\cite{Cai2005}              &   8,293                                        & 18,933      & 785 (85\%)        & 65   \\
PenDigits                       &   10,992                                       & 16         & \hspace{1.5mm}16         & 10   \\
Covtype~\cite{Alimoglu1996}     &   581,012                                      & 54         & \hspace{1.5mm}54         & 7    \\
PokerHand-1~\cite{Blackard1999}   &   971,329                                    & 10         & \hspace{1.5mm}10         & 3   \\
PokerHand-2~\cite{Blackard1999}   &   1,000,000                                   & 10         & \hspace{1.5mm}10         & 10   \\
\bottomrule
\end{tabular}
\end{center}
\end{table}

\subsection{Baseline Algorithms and Evaluation Metrics}
\label{sec4.2}

Spectral clustering and kernel-based clustering methods are popular to cope with linearly inseparable data. Some studies~\cite{Filippone2008} have established the equivalence between them. In the experiments, we compare the proposed methods with four scalable spectral clustering algorithms (KASP~\cite{Yan2009}, Nystr\"{o}m approximation based spectral clustering~\cite{Fowlkes2004,Chen2011a}, LSC~\cite{Chen2011}, and SEC~\cite{Nie2011}) and one scalable kernel-based clustering approach (AKK~\cite{Chitta2011}). Moreover, we report the results of the k-means clustering algorithm~\cite{Cai2011kmeans} as a baseline. Besides our own implementation, we also quote some results directly from the literature.

We investigate the performance of two variants of Nystr\"{o}m-based methods and LSC, denoted as Nystr\"{o}m, Nystr\"{o}m-Orth, LSC\_R, and LSC\_K. The affinity matrix of Nystr\"{o}m-Orth is orthogonal, whereas that of Nystr\"{o}m is not. SEC obtains the results by performing k-means in the embedding space. All algorithms are implemented in MATLAB. The used data sets and the codes of our algorithms can be downloaded at \textcolor{blue}{\url{www.machineilab.org/users/pengxi/}}.

\begin{figure*}[t]
\centering
\subfigure [The influences of the parameter $\lambda$ of SSSC, where $\delta=10^{-3}$.]{\label{fig:5.a}\includegraphics[width=0.43\textwidth]{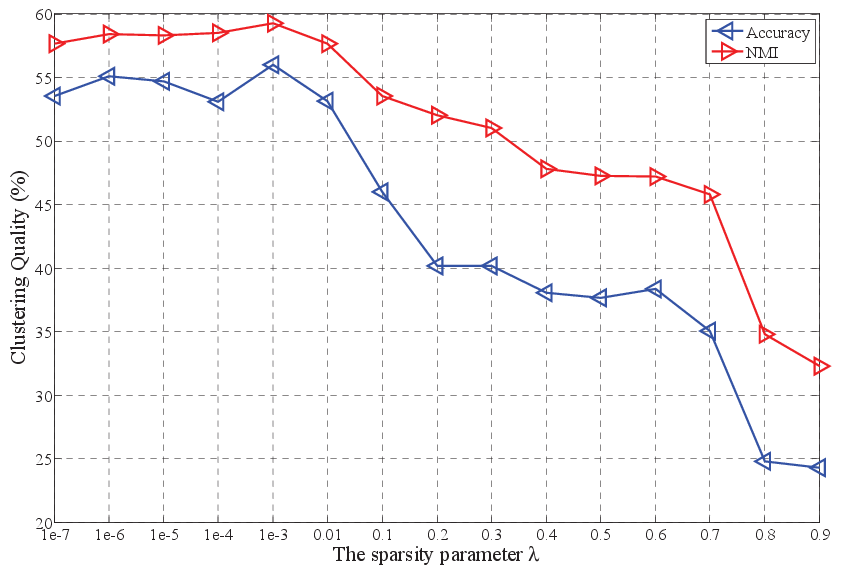}}\hspace{0.6cm}
\subfigure [The influences of the parameter $\delta$ of SSSC, where $\lambda=10^{-3}$.]{\label{fig:5.b}\includegraphics[width=0.42\textwidth]{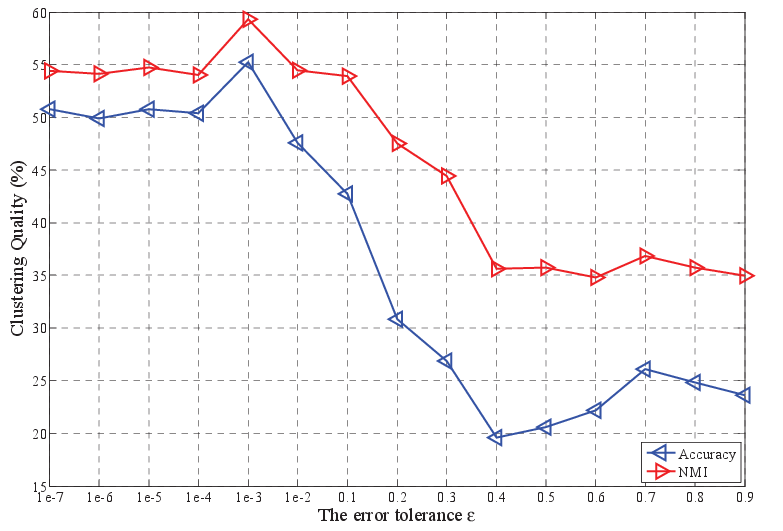}}\\
\subfigure [The influences of the parameter $\lambda$ of SLRR.]{\label{fig:5.c}\includegraphics[width=0.43\textwidth]{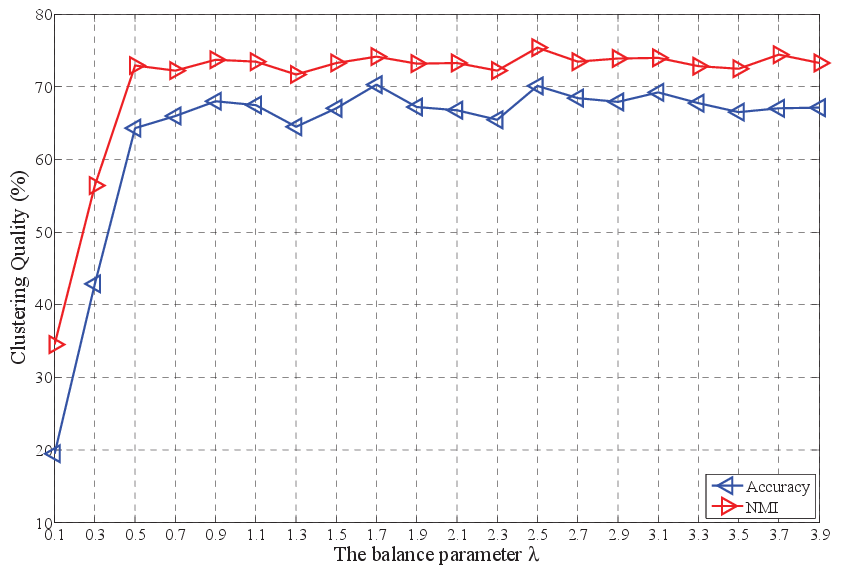}}\hspace{0.6cm}
\subfigure [The influences of the parameter $\lambda$ of SLSR.]{\label{fig:5.d}\includegraphics[width=0.42\textwidth]{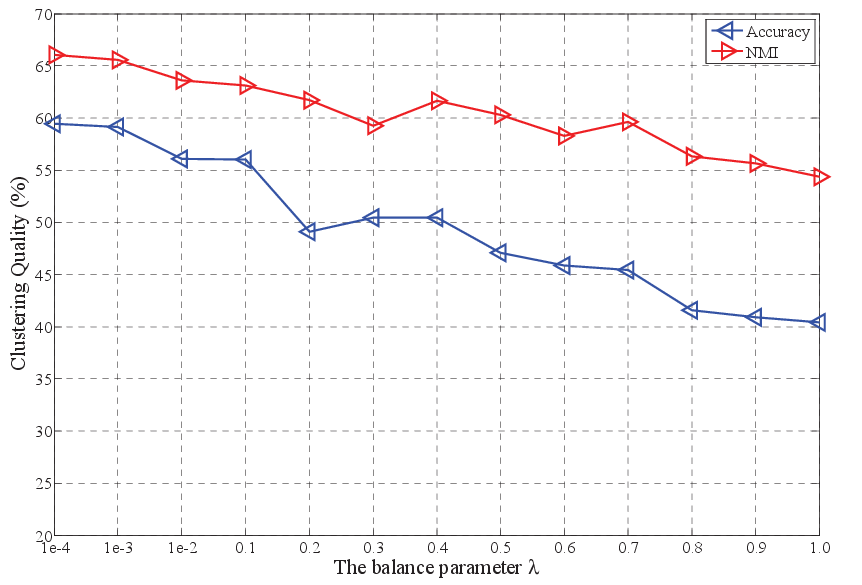}}
\caption{The influence of the parameters. A half of images (1212) are chosen from ExYaleB as in-sample data and the rest are used as out-of-sample data. The x-coordinate denotes the values of the parameters, and the y-coordinate corresponds to the clustering quality (\emph{Accuracy} and \emph{NMI}).}
\label{fig:5}
\end{figure*}

The evaluated algorithms take two approaches to find in-sample data. Specifically, SSSC, SLRR, SLSR, Nystr\"{o}m, Nystr\"{o}m\_Orth, LSC\_R, SEC and AKK identify in-sample data by performing uniform random sampling method, whereas KASP and LSC\_K adopt the k-means clustering method. To avoid the disparity in data partitions, we pre-partition each data set into two parts, in-sample data and out-of-sample data. After that, we run different algorithms  run over these data partitions.

We measure the clustering quality using \emph{Accuracy}~\cite{Zhao2001} and Normalized Mutual Information (\emph{NMI})~\cite{Cai2005} between the produced clusters and the ground truth categories. The \emph{Accuracy} or \emph{NMI} of 1 indicates perfect matching with the true subspace distribution, whereas 0 indicates totally mismatch.

%
%
%
%

To be consistent with the previous works~\cite{Elhamifar2012,Liu2012}, we tune the parameters of all the evaluated methods to achieve the highest \emph{Accuracy}. For SSSC, we adopted the Homotopy optimizer to solve the $\ell_1$-minimization problem. The optimizer has two user-specified parameters, sparsity parameter $\lambda$ and error tolerance parameter $\delta$. We tuned the parameters in the range of $\lambda=(10^{-7},10^{-6},10^{-5})$ and $\delta=(10^{-3},10^{-2},10^{-1})$. For SLRR and SLSR, the value of $\lambda$ is chosen as shown in~\figurename~\ref{fig:5}. Referring to the parameter setting in
~\cite{Chen2011a,Yan2009,Chen2011,Nie2011,Chitta2011},
the parameter $\tau$ of KASP and Nystr\"{o}m was set as $[0.1, 1]$ with an interval of 0.1 and $[2, 20]$ with an interval of 1; the parameter $\sigma$ of AKK ranges from $[0.1, 1]$ with an interval of $0.1$; SEC has three user-specified parameters, \emph{i.e.}, the size of neighborhood $r$, balanced parameters $\mu$ and $\gamma$. We set $\gamma=1$, $\mu=[10^{-9},10^{-6},10^{-3},10^{0},10^{+3},10^{+6},10^{+9},10^{+12},10^{+15}]$, and $r$ from $2$ to $20$. Moreover, the same value range of $r$ was used for KASP and LSC.

Following the common benchmarking procedures, we run  each algorithm five times on each data set and report the final results by the mean and standard deviation of the \textit{Accuracy} (\textit{NMI}) and the mean of time costs.

\subsection{The Influence of Parameters}
\label{sec4.3}

SSSC uses $\lambda>0$ to control the sparsity of the representation and $\epsilon>0$ to measure the reconstruction errors. SLRR uses $\lambda>0$ to balance different parts in the objective function and SLSR utilizes $\lambda>0$ to avoid overfitting. The choice of these parameters depends on the data distribution.

\figurename~\ref{fig:5} shows the results of SSSC, SLRR, and SLSR with different parameter values. When $\lambda$ or $\epsilon$ of SSSC is assigned with a small positive value (from $10^{-7}$ to $0.01$), it achieves a good performance. When the parameters are assigned with a big value, the performance of SSSC is degraded. For SLRR, while $\lambda$ ranges from 0.5 to 3.9, its \emph{Accuracy} and \emph{NMI} are almost
unchanged. SLSR performs worse with increasing $\lambda$. This verifies our claim that a small $\lambda$ is preferable to the clean data set.

\subsection{The Influence of In-sample Data Size}
\label{sec4.4}

\begin{figure*}[t]
\centering
\subfigure [\emph{Accuracy} versus the varying in-sample data size]{\label{fig:6.a}\includegraphics[width=0.42\textwidth]{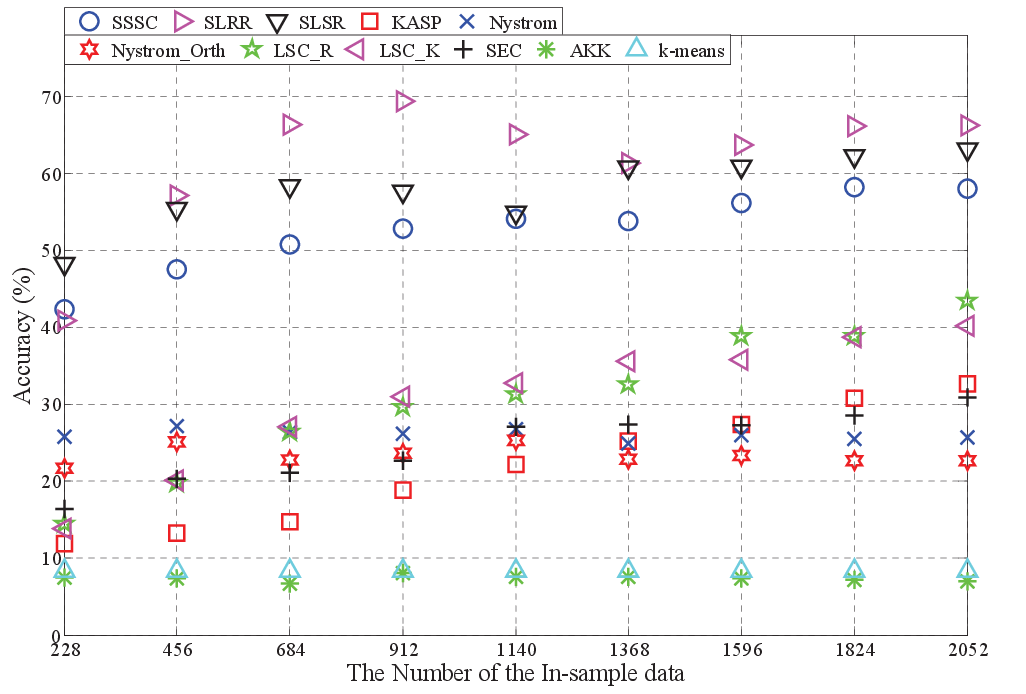}}\hspace{1cm}
\subfigure [\emph{NMI} versus the varying in-sample data size]{\label{fig:6.b}\includegraphics[width=0.42\textwidth]{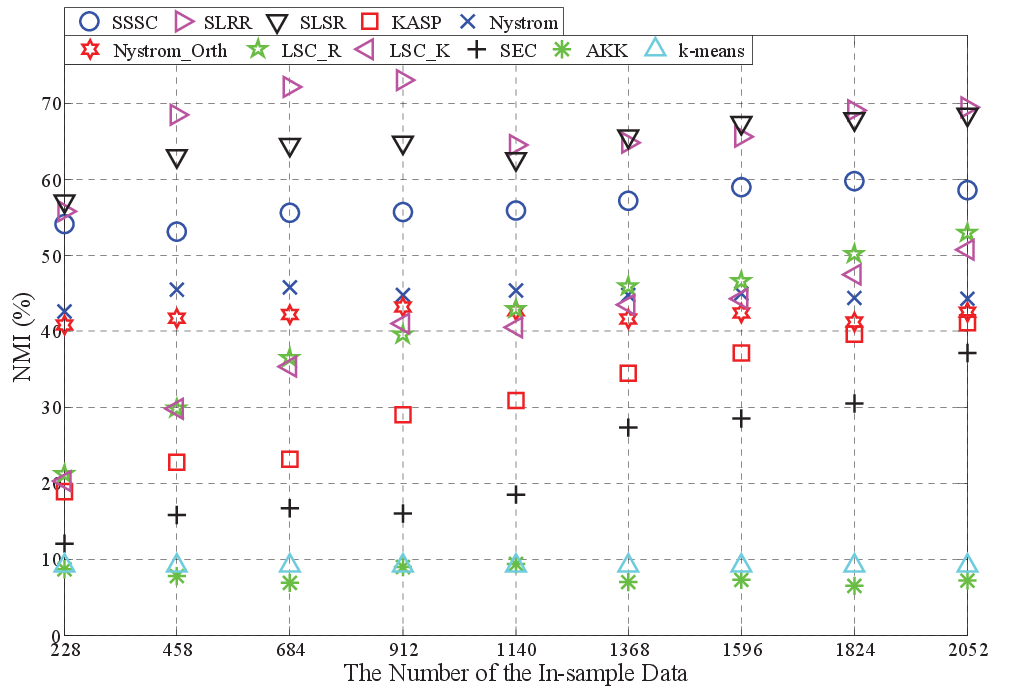}}
\caption{Clustering quality of the competing algorithms on the Extended Yale database B. The x-coordinate denotes in-sample data size and the y-coordinate denotes the clustering quality (\emph{Accuracy} or \emph{NMI}).}
\label{fig:6}
\end{figure*}

 To study the influences of in-sample data size $p$, we perform experiments on ExYaleB by setting $p=38\times \tilde{p}$, where $\tilde{p}$ denotes the sample size per subject and it increases from 6 to 54 with an interval of 6.~\figurename~\ref{fig:6} reports the result, from which we have the following observations:

\begin{itemize}
  \item Except SEC and AKK, all the scalable clustering methods outperform the k-means method in \emph{Accuracy} and \emph{NMI}. SSSC, SLRR, and SLSR are superior to the other investigated approaches by a considerable performance margin. For example, SLRR achieves 15.1\% gain in \emph{Accuracy} and 13.1\% gain in \emph{NMI} over the best baseline algorithm (Nystr\"{o}m) when $p=228$.
  \item In most cases, all the algorithms except Nystr\"{o}m and Nystr\"{o}m\_Orth perform better with increasing $p$. The possible reason for this result is that Nystr\"{o}m and Nystr\"{o}m\_Orth speed up the clustering process by reducing the size of affinity matrix rather than data size.
  \item The accuracy of SLRR decreased when $p$ increased from $912$ to $1368$. The result seems inconsistent with the common sense that more data tend to bring better performance. This result can be attributed to the characteristic of SLRR, i.e., SLRR is based on low rank representation that incorporates  the relations among different subspaces. Increasing $p$ would result in more intersections among different subspaces and weaken the discrimination of model. To obtain an optimal $p$, some model selection methods such as $M$-estimator~\cite{negahban2011estimation} could be used.
   \end{itemize}

\subsection{Clustering on Small Scale Data}
\label{sec4.5}

We carry out the experiments on three facial data sets, \emph{i.e.}, AR, ExYaleB, and LFW. Moreover, we investigate the performance of our methods when the classifiers (\ref{equ:9}) and (\ref{equ:10}) are used to group out-of-sample data. In the experiments, we fix $\epsilon=10^{-3}$ for SSSC and SSC.

From Table~\ref{tab:4}, we can find that
\begin{itemize}
  \item Our framework successfully makes SSC, LRR, and LSR feasible to group out-of-sample data with acceptable loss in clustering quality. For example, the \emph{Accuracy} of SSC on AR data set is 9.73\% higher than that of SSSC, whereas the time cost of SSC is about three times that of SSSC. With the increase of data size, SSC, LRR, and LSR will fail to get the results, whereas SSSC, SLRR, SLSR can get the results with an acceptable time cost.
  \item Compared with the other scalable methods (\emph{i.e.}, KASP, Nystr\"{o}m, Nystr\"{o}m\_Orth, LSC\_R, LSC\_K, SEC, and AKK), SSSC, SLRR, and SLSR find an elegant balance between the clustering quality and the time costs. Although SSSC, SLRR, and SLSR are not the fastest, they achieve the best results.
  \item SLRR performs better than SSSC in the tests. The possible reason is that the low rank representation could capture the structure among different categories, whereas sparse representation cannot, as pointed out in~\cite{Liu2012FRR}. Moreover, the regularized residual based classifier (\ref{equ:10}) perform slightly better than the non-regularized residual based classification method (\ref{equ:9}).
  \item Nie \emph{et\ al.}~\cite{Nie2011} investigated the performance of SEC on ExYaleB. The highest \emph{Accuracy} of SEC is about 42.8\% in their tests, comparing with 22.02\% in our experiment. The potential reason for the performance difference is that they adopted spectral rotation to get the cluster membership, whereas we use the k-means clustering method. Note that, the best result (42.8\%) of SEC reported in their work is remarkably lower than the results achieved by SSSC ($55.5\pm1.26$), SLRR ($68.9\pm1.19$), and SLSR ($58.9\pm1.45$).
\end{itemize}

\begin{table*}[t]
\caption{Performance comparison (mean$\pm$std) among different algorithms over three popular facial data sets. }
\label{tab:4}
\centering
\begin{footnotesize}
\begin{tabular}{l llr| llr| llr}
\toprule
Data sets & \multicolumn{3}{c|}{AR ($p=700$)} & \multicolumn{3}{c|}{ExYaleB ($p=1212$)} & \multicolumn{3}{c}{LFW ($p=1000$)} \\
\midrule
Algorithm & \multicolumn{1}{c}{Accuracy (\%)} & \multicolumn{1}{c}{NMI (\%)} & \multicolumn{1}{c|}{Time} & \multicolumn{1}{c}{Accuracy (\%)} & \multicolumn{1}{c}{NMI (\%)} & \multicolumn{1}{c|}{Time}& \multicolumn{1}{c}{Accuracy (\%)} & \multicolumn{1}{c}{NMI (\%)} & \multicolumn{1}{c}{Time} \\
\hline
SSSC	 & 	60.4$\pm$1.74($10^{-5}$) &80.8$\pm$0.99 & 142.2  & 55.5$\pm$1.26($10^{-5}$) &60.3$\pm$0.29 & 128.0  & 27.6$\pm$0.51($10^{-6}$) &43.7$\pm$0.12 & 184.0 \\
SLRR & 70.6$\pm$1.50($3.1$) &87.3$\pm$0.44 & 40.1  & \textbf{68.9$\pm$1.19}($2.9$) &\textbf{74.0$\pm$0.60} & 26.8  & \textbf{30.4$\pm$0.46}($0.7$) &\textbf{44.8$\pm$0.19} & 228.4 \\
SLSR & \textbf{78.7$\pm$1.42}($10^{-2}$) & \textbf{89.6$\pm$0.40} & 32.4  & 58.9$\pm$1.45($10^{-4}$) &65.2$\pm$0.61 & 21.4  & 28.5$\pm$0.32($0.7$) &43.8$\pm$0.32 & 213.4 \\
KASP~\cite{Yan2009} & 32.5$\pm$0.55($0.1$) &63.6$\pm$0.57 & 134.8  & 20.6$\pm$1.28($8$) &31.3$\pm$0.93 & 37.8  & 25.1$\pm$0.93($7$) &42.7$\pm$0.40 & 251.4 \\
Nystr\"{o}m~\cite{Chen2011a} & 62.2$\pm$1.71($2$) &82.1$\pm$1.16 & 2.3  & 20.7$\pm$1.16($12$) &39.7$\pm$0.63 & 8.2  & 26.6$\pm$0.81($0.6$) &42.0$\pm$0.33 & 3.2 \\
Nystr\"{o}m\_Orth & 57.5$\pm$3.55($0.9$) &79.1$\pm$1.80 & 13.7  & 21.4$\pm$1.50($3$) &40.3$\pm$1.01 & 60.9  & 26.7$\pm$0.86($0.5$) &41.0$\pm$0.37 & 11.3 \\
LSC\_R~\cite{Chen2011} & 31.1$\pm$0.71($4$) &61.3$\pm$0.52 & 1.7  & 32.3$\pm$0.91($2$) &43.7$\pm$0.34 & 7.3  & 25.9$\pm$0.48($3$) &41.5$\pm$0.28 & 3.9 \\
LSC\_K~\cite{Chen2011} & 32.9$\pm$0.79($4$) &62.9$\pm$0.50 & 2.2  & 31.2$\pm$2.07($2$) &42.1$\pm$1.33 & 8.3  & 22.0$\pm$0.50($10$) &41.6$\pm$0.28 & 5.1 \\
SEC~\cite{Nie2011} & 25.9$\pm$1.81($10^{+9},8$) &41.1$\pm$1.60 & 1.7  & 22.0$\pm$1.68($10^{-9},1$) &39.4$\pm$1.85 & 10.3  & 25.2$\pm$1.68($10^{+12},4$) &40.4$\pm$1.49 & 2.3 \\
AKK~\cite{Chitta2011} & 22.0$\pm$1.28($0.2$) &52.0$\pm$1.09 & 0.8  & \hspace{1.4mm}6.8$\pm$0.48($0.4$) &\hspace{1.4mm}5.5$\pm$0.82 & 3.0  & 16.0$\pm$0.99($0.3$) &34.7$\pm$0.81 & 2.7 \\
SSSC2 & 58.3$\pm$1.38($10^{-5}$) &79.6$\pm$0.49 & 79.6  & 57.8$\pm$1.21($10^{-5}$) &62.3$\pm$0.60 & 65.0  & 26.5$\pm$0.22($10^{-6}$) &43.2$\pm$0.11 & 212.7 \\
SLRR2 & 69.1$\pm$2.50($3.1$) &86.2$\pm$0.77 & 39.8  & 71.8$\pm$0.91($2.9$) &77.3$\pm$0.45 & 30.1  & 29.1$\pm$0.49($0.7$) &43.9$\pm$0.08 & 321.6 \\
SLSR2 & 77.6$\pm$1.30($10^{-2}$) &88.7$\pm$0.55 & 30.1  & 61.2$\pm$1.35($10^{-4}$) &67.3$\pm$0.90 & 23.6  & 28.8$\pm$0.39($0.7$) &43.6$\pm$0.15 & 232.4 \\
\hline
k-means~\cite{Cai2011kmeans} & 29.1$\pm$0.59(-) &58.4$\pm$0.43 & 18.8  &\hspace{1.4mm}8.4$\pm$0.50(-) &\hspace{1.4mm}9.9$\pm$0.72 & 50.2  & 19.4$\pm$0.56(-) &37.3$\pm$0.27 & 87.7 \\
SSC~\cite{Elhamifar2012} & 70.1$\pm$1.85($10^{-7}$) &86.4$\pm$0.73 & 361.4  & 59.0$\pm$0.91($10^{-3}$) &65.1$\pm$0.34 & 344.9  & 31.6$\pm$0.64($10^{-5}$) &47.5$\pm$0.24 & 804.0 \\
LRR~\cite{Liu2012} & 78.6$\pm$0.02($1$) &89.3$\pm$0.59 & 152.9  & \textbf{73.7$\pm$0.01}($2.1$) &\textbf{78.5$\pm$0.46} & 46.9  & 36.4$\pm$0.02($3.2$) &51.3$\pm$0.36 & 623.1 \\
LSR~\cite{Lu2012} & \textbf{81.4$\pm$1.77}($10^{-2}$) &\textbf{91.4$\pm$0.60} & 104.8  & 68.7$\pm$2.11($0.2$) &72.9$\pm$1.58 & 89.6  & \textbf{37.9$\pm$0.66}($0.9$) &\textbf{54.1$\pm$0.23} & 243.9 \\
\bottomrule
\end{tabular}
\end{footnotesize}
\footnotesize \begin{flushleft}Note: $p$ denotes in-sample data size. The number in the parenthesis are the tuned parameters. SSC, LRR, LSR, and the k-means method cannot handle out-of-sample data. Thus, the results of these four methods are achieved by directly performing them on the whole data set. SSSC, SLRR, and SLSR assign out-of-sample data to the nearest subspace which has minimal residual (\emph{i.e.}, eq.(\ref{equ:10})), whereas SSSC2, SLRR2, and SLSR2 get the results using  eq.(\ref{equ:9}). The \textbf{bold number} indicate the best performance.\end{flushleft}
\end{table*}

\subsection{Clustering on Medium Scale Data}
\label{sec4.6}

 This section investigates the performance of our methods on MPIE (facial images), RCV (documental corpus), and PenDigits (handwritten digital data). The tuned $\epsilon$ of SSSC are $10^{-4}$, $0.1$, and $0.01$, respectively. Table~\ref{tab:5} reports the clustering quality and the time cost (seconds) of the tested methods, from which we can find that

\begin{table*}[t]
\caption{Performance comparison among different algorithms on three medium-sized data sets, \emph{i.e.}, MPIE, RCV, and PenDigits.}
\label{tab:5}
\centering
\begin{footnotesize}
\begin{tabular}{l llr| llr| llr}
\toprule
Data sets & \multicolumn{3}{c|}{MPIE ($p=1000$)} & \multicolumn{3}{c|}{PenDigits ($p=1000$)} & \multicolumn{3}{c}{RCV ($p=2000$)} \\
\midrule
Algorithm & \multicolumn{1}{c}{Accuracy (\%)} & \multicolumn{1}{c}{NMI (\%)} & \multicolumn{1}{c|}{Time} & \multicolumn{1}{c}{Accuracy (\%)} & \multicolumn{1}{c}{NMI (\%)} & \multicolumn{1}{c|}{Time}& \multicolumn{1}{c}{Accuracy (\%)} & \multicolumn{1}{c}{NMI (\%)} & \multicolumn{1}{c}{Time} \\
\hline
SSSC	 & 57.6$\pm$0.97($10^{-6}$) &\textbf{79.4$\pm$0.40} &  432.1  &  \textbf{80.0$\pm$1.31}($10^{-7}$) &71.3$\pm$0.11 &  17.0  &  19.6$\pm$1.34($10^{-7}$) &29.8$\pm$0.58 &  840.6 \\
SLRR & \textbf{60.7$\pm$0.62}($2.30 $) &78.9$\pm$0.34 &  340.4  &  74.8$\pm$0.92($0.30 $) &67.6$\pm$0.00 &  10.4  &  \textbf{49.1$\pm$0.11}($3.10 $) &31.3$\pm$0.26 &  499.6 \\
SLSR & 59.0$\pm$0.58($0.60 $) &79.5$\pm$0.49 &  355.4  &  78.4$\pm$0.81($1.00 $) &69.6$\pm$0.01 &  8.9  &  11.2$\pm$0.41($0.60 $) &18.3$\pm$1.22 &  95.2 \\
KASP~\cite{Yan2009} & 16.6$\pm$0.53($0.1$) &57.0$\pm$0.28 &  1479.8  &  73.1$\pm$6.37($4$) &\textbf{75.5$\pm$3.39} &  12.5  &  19.0$\pm$0.64($0.1$) &26.7$\pm$0.33 &  198.8 \\
Nystr\"{o}m~\cite{Chen2011a} & 47.1$\pm$1.46($0.7$) &77.2$\pm$.0.88 &  15.3  &  66.7$\pm$6.93($0.4$) &65.4$\pm$2.70 &  35.9  &  15.9$\pm$1.10($0.4$) &27.7$\pm$0.37 &  27.1 \\
Nystr\"{o}m\_Orth & 50.3$\pm$.2.38($0.7$) &78.1$\pm$1.62 &  64.8  &  67.3$\pm$5.66($3$) &64.8$\pm$2.67 &  6.2  &  19.8$\pm$0.53($0.1$) &23.7$\pm$0.39 &  3401.3 \\
LSC\_R~\cite{Chen2011} & 18.1$\pm$0.11($2$) &54.5$\pm$0.25 &  62.1  &  77.7$\pm$3.18($15$) &74.9$\pm$2.61 &  5.6  &  15.4$\pm$0.15($2$) &22.2$\pm$0.15 &  8.9 \\
LSC\_K~\cite{Chen2011} & 17.5$\pm$0.37($3$) &56.1$\pm$0.46 &  65.7  &  79.9$\pm$2.73($11$) &76.4$\pm$0.58 &  7.9  &  22.0$\pm$1.83($2$) &\textbf{34.5$\pm$0.43} &  17.7 \\
SEC~\cite{Nie2011} & 13.2$\pm$0.39($10^{-3},9$) &44.1$\pm$0.43 &  27.2  &  75.3$\pm$4.20($10^{-9},4$) &70.3$\pm$2.43 &  11.8  &  14.8$\pm$0.67($10^{-6},3$) &26.3$\pm$0.52 &  19.9 \\
AKK~\cite{Chitta2011} & 10.4$\pm$0.19($0.1$) &38.7$\pm$0.66 &  24.6  &  69.0$\pm$4.64($0.01$) &66.9$\pm$1.63 &  6.2  &  18.3$\pm$0.62($0.2$) &31.6$\pm$0.30 &  27.9 \\
k-means~\cite{Cai2011kmeans} & 14.5$\pm$0.36(-) &53.2$\pm$0.26 &  268.5  &  77.0$\pm$0.13(-) &69.2$\pm$0.02 &  23.7  &  19.3$\pm$1.10(-) &23.8$\pm$0.52 &  256.8 \\
\bottomrule
\end{tabular}
\end{footnotesize}
\end{table*}

\begin{itemize}
  \item Our methods outperform the other scalable methods. For example, SLRR achieves a 10.4\% gain in \emph{Accuracy} on MPIE over the best competing algorithm (Nystr\"{o}m\_Orth), and the gains achieved by SSSC and SLSR are about 7.3\% and 8.6\%, respectively.
  \item The running time is a weakness of SSSC, SLRR, and SLSR even though they are more efficient than the original approaches. We have found that most of the time was consumed to handle in-sample data. For example, SSSC takes 840.6 seconds to cluster in-sample data and 220.63 seconds to handle out-of-sample data in the case of RCV. Since in-sample data clustering is an offline process, we assume that our algorithms are more competitive in large scale setting as shown in Section~\ref{sec4.7}.
  \item In most cases, LSC\_K outperforms LSC\_R with a little improvement, which verifies the claim~\cite{Kvalseth1987} that the complex sampling techniques actually cannot produce a better result than the random sampling method.
  \item \cite{Chen2011} also investigated the \emph{Accuracy} of LSC\_R, LSC\_K, Nystr\"{o}m\_Orth, and KASP on the PenDigits data set. The highest \emph{Accuracy} of these algorithms are 79.0\%, 79.3\%, 73.9\% and 72.5\%, which is close to the results achieved in our experiments (\emph{i.e.}, 77.7\%, 79.9\%, 67.3\% and 73.1\%).
  \end{itemize}

\subsection{Clustering on Large Scale Data}
\label{sec4.7}

Table~\ref{tab:6} reports the performance of our algorithms on three large scale data sets. For each data set, 1000 samples are selected as in-sample data, and the remaining samples are used as out-of-sample data. We assign $\epsilon=0.2$ to SSSC on Covtype and PokerHand-2 and fix $\epsilon=0.1$ in the case of PokerHand-1. We have the following observations:

\begin{table*}[t]
\caption{Performance comparison among different algorithms over three large scale data sets, \emph{i.e.},
Covtype ($n=581,012$), PokerHand-1 ($n=971,329$), and PokerHand-2 ($n=1,000,000$). }
\label{tab:6}
\centering
\begin{footnotesize}
\begin{tabular}{l llr| llr| llr}
\toprule
Data sets & \multicolumn{3}{c|}{Covtype ($p=1000$)} & \multicolumn{3}{c|}{PokerHand-1 ($p=1000$)} & \multicolumn{3}{c}{PokerHand-2 ($p=1000$)} \\
\midrule
Algorithm & \multicolumn{1}{c}{Accuracy (\%)} & \multicolumn{1}{c}{NMI (\%)} & \multicolumn{1}{c|}{Time} & \multicolumn{1}{c}{Accuracy (\%)} & \multicolumn{1}{c}{NMI (\%)} & \multicolumn{1}{c|}{Time}& \multicolumn{1}{c}{Accuracy (\%)} & \multicolumn{1}{c}{NMI (\%)} & \multicolumn{1}{c}{Time} \\
\hline
SSSC	 & 	\textbf{28.6$\pm$0.00}($	10^{-5}$) &	5.3$\pm$0.00	 &  	325.5 	 &  	\textbf{51.6$\pm$0.00}($	10^{-7}$) &	\textbf{0.3$\pm$0.00}	 &  	267.7 	 &  	\textbf{17.6$\pm$0.00}($	10^{-5}$) &	0.1$\pm$0.10	 &  	474.1 	\\
SLRR	 & 	27.1$\pm$0.03($	0.10 $) &	3.6$\pm$0.02	 &  	240.9 	 &  	37.8$\pm$0.00($	0.10 $) &	0.1$\pm$0.00	 &  	166.9 	 &  	16.0$\pm$0.00	($	0.10 $) &	0.1$\pm$0.00	 &  	317.7 	\\
SLSR	 & 	26.5$\pm$0.00($	0.01 $) &	\textbf{7.2$\pm$0.00}	 &  	268.8 	 &  	37.0$\pm$0.00($0.10 $) &0.0$\pm$0.00 &  167.8  &  15.8$\pm$0.01($0.10 $) &0.1$\pm$0.00 &  494.2 \\
KASP~\cite{Yan2009} & 23.9$\pm$1.93($3$) &3.5$\pm$0.19 &  1314.5  &  34.7$\pm$0.93($0.3$) &0.0$\pm$0.00 &  5497.1  &  11.3$\pm$0.32($3$) &0.1$\pm$0.04 &  7049.9 \\
Nystr\"{o}m~\cite{Chen2011a} & 24.0$\pm$0.59($0.1$) &3.8$\pm$0.03 &  40.6  &  47.9$\pm$0.02($0.2$) &0.2$\pm$0.01 &  61.4  &  12.9$\pm$0.27($0.2$) &\textbf{0.2$\pm$0.04} &  205.7 \\
Nystr\"{o}m\_Orth & 23.3$\pm$0.67($0.1$) &3.8$\pm$0.16 &  351.6  &  35.8$\pm$0.33($20$) &0.1$\pm$0.00 &  204.4  &  15.6$\pm$2.89($17$) &0.1$\pm$0.02 &  205.7 \\
LSC\_R~\cite{Chen2011} & 22.0$\pm$0.47($2$) &3.8$\pm$0.06 &  154.5  &  34.9$\pm$0.01($8$) &0.0$\pm$0.00 &  1891.0  &  12.6$\pm$0.17($5$) &0.0$\pm$0.04 &  1936.8 \\
LSC\_K~\cite{Chen2011} & 22.0$\pm$0.52($4$) &3.6$\pm$0.10 &  1155.4  &  32.4$\pm$1.03($2$) &0.0$\pm$0.00 &  8765.5  &  13.8$\pm$0.51($3$) &0.1$\pm$0.02 &  8829.0 \\
SEC~\cite{Nie2011} & 21.1$\pm$0.01($1,4$) &3.6$\pm$0.00 &  64.9  &  36.6$\pm$0.00($10^{-9},3$) &0.1$\pm$0.00 &  81.4  &  10.5$\pm$0.06($10^{-3},4$) &0.1$\pm$.0.01 &  130.2 \\
AKK~\cite{Chitta2011} & 22.8$\pm$1.63($1$) &3.8$\pm$0.08 &  344.2  &  35.9$\pm$0.04($0.1$) &0.1$\pm$0.00 &  1039.3  &  10.5$\pm$0.06($0.01$) &0.0$\pm$0.01 &  2882.5 \\
k-means~\cite{Cai2011kmeans} & 20.8$\pm$0.00(-) &3.7$\pm$0.00 &  4895.7  &  36.0$\pm$0.01(-) &0.1$\pm$0.00 &  4760.4  &  10.4$\pm$0.06(-) &0.0$\pm$0.01 &  7188.8 \\
\bottomrule
\end{tabular}
\end{footnotesize}
\end{table*}

\begin{itemize}
  \item SSSC, SLRR, and SLSR outperform the other approaches in all the tests. For example, the \emph{Accuracy} of SSSC is at least 4.7\% higher than the other tested methods on Covtype. On PokerHand-1 and PokerHand-2, the gains are 3.7\% and 2.1\%, respectively.
  \item The \emph{NMI} achieved by all the tested methods are close to 0. This shows that the metric \emph{NMI} failed to distinct the performance of the evaluated algorithms.
  \item In~\cite{Chen2011}, the highest \emph{Accuracy} on Covtype achieved by LSC\_R, LSC\_K, Nystr\"{o}m\_Orth and KASP are 24.7\%, 25.5\%, 22.3\% and 22.4\%, respectively. In our experiments, the \emph{Accuracy} of these four algorithms are 22.0\%, 22.0\%, 23.3\% and 23.9\%, respectively. The possible reason may attribute to the subtle engineering details, \emph{e.g.}, the in-sample and out-of-sample data partitions.
  \item With the increase of data size, our methods demonstrate a good balance between the running time and the clustering quality. Moreover, the used memory of our methods only depends on in-sample data size, which makes our methods are very competitive in large scale setting.
\end{itemize}

In summary, we can conclude that the three new methods outperform the competing algorithms in all the tests.
In particular, SSSC is more advantageous on large scale data sets  (\emph{e.g.}, Covtype and PokerHand), while SLRR outperforms on high-dimensional data clustering problems (\emph{e.g.}, facial images and documental corpus). SLSR can achieve comparable clustering performance with SSSC and SLRR, but has higher computational efficiency than the latter.
\section{Conclusion}
\label{sec5}

In this paper, we proposed a general framework to solve the large-scale and the out-of-sample clustering problems for representation-based subspace clustering. Under our framework, we further presented three scalable methods, \emph{i.e.}, SSSC, SLRR, and SLSR, which largely reduce the computational complexity of the original methods while preserving a good performance. We proved that the performance of our method only depends on the latent structure of the data set and is independent of the sampling rate. Moreover, we proposed a novel method to analyze the error bounds of the nearest subspace classifier in terms of binary case and applied it to SRC. Both theoretical and experimental results show the effectiveness of our methods in large scale clustering.

The work may be extended or improved from the following aspects. First, the proposed framework is based on the assumption that out-of-sample data can be represented by in-sample data. Hence, the method may fail to handle the out-of-sample datum when it comes from a new subspaces that does not emerge from in-sample data. It is worth to explore how to overcome this problem in future. Second, the proposed error analysis method only considers the binary case (i.e., $k=2$). It is more practical but challenging to explore the error analysis method w.r.t. $k>2$.


%

%

\section*{Acknowledgment}
The authors would like to thank the anonymous reviewers for their valuable comments and suggestions to improve the quality of this paper. This work was supported by National Nature
Science Foundation of China under grant No.61432012 and No. 61322203.

\ifCLASSOPTIONcaptionsoff
  \newpage
\fi

\bibliographystyle{IEEEtran}
\bibliography{IEEEabrv,SRSC}

\begin{thebibliography}{10}
\providecommand{\url}[1]{#1}
\csname url@samestyle\endcsname
\providecommand{\newblock}{\relax}
\providecommand{\bibinfo}[2]{#2}
\providecommand{\BIBentrySTDinterwordspacing}{\spaceskip=0pt\relax}
\providecommand{\BIBentryALTinterwordstretchfactor}{4}
\providecommand{\BIBentryALTinterwordspacing}{\spaceskip=\fontdimen2\font plus
\BIBentryALTinterwordstretchfactor\fontdimen3\font minus
  \fontdimen4\font\relax}
\providecommand{\BIBforeignlanguage}[2]{{%
\expandafter\ifx\csname l@#1\endcsname\relax
\typeout{** WARNING: IEEEtran.bst: No hyphenation pattern has been}%
\typeout{** loaded for the language `#1'. Using the pattern for}%
\typeout{** the default language instead.}%
\else
\language=\csname l@#1\endcsname
\fi
#2}}
\providecommand{\BIBdecl}{\relax}
\BIBdecl

\bibitem{Yu2015Generalized}
Z.~Yu, W.~Liu, W.~Liu, X.~Peng, Z.~Hui, and B.~V. Kumar, ``Generalized
  transitive distance with minimum spanning random forest,'' in \emph{Proc. of
  24th Int. Joint Conf. on Artif. Intell.}, Buenos Aires, Argentina, Jul. 2015,
  pp. 2205--2211.

\bibitem{Muller2001}
K.~Muller, S.~Mika, G.~Ratsch, K.~Tsuda, and B.~Scholkopf, ``An introduction to
  kernel-based learning algorithms,'' \emph{IEEE Trans. Neural. Netw.},
  vol.~12, no.~2, pp. 181--201, 2001.

\bibitem{Vidal2011}
R.~Vidal, ``Subspace clustering,'' \emph{IEEE Signal Proc. Mag.}, vol.~28,
  no.~2, pp. 52--68, 2011.

\bibitem{Ma2007}
Y.~Ma, H.~Derksen, W.~Hong, and J.~Wright, ``Segmentation of multivariate mixed
  data via lossy data coding and compression,'' \emph{IEEE Trans. Pattern Anal.
  Mach. Intell.}, vol.~29, no.~9, pp. 1546--1562, 2007.

\bibitem{Rao2008}
S.~Rao, R.~Tron, R.~Vidal, and Y.~Ma, ``Motion segmentation via robust subspace
  separation in the presence of outlying, incomplete, or corrupted
  trajectories,'' in \emph{Proc. of 21th IEEE Conf. Comput. Vis. and Pattern
  Recognit.}, Anchorage, AL, Jun. 2008, pp. 1--8.

\bibitem{Ng2002}
A.~Y. Ng, M.~I. Jordan, and Y.~Weiss, ``On spectral clustering: Analysis and an
  algorithm,'' in \emph{Proc. of 14th Adv. in Neural Inf. Process. Syst.},
  Vancouver, Canada, Dec. 2001, pp. 849--856.

\bibitem{Hou2014}
C.~Hou, F.~Nie, D.~Yi, and D.~Tao, ``Discriminative embedded clustering: A
  framework for grouping high-dimensional data,'' \emph{IEEE Trans. Neural.
  Netw. Learn. Syst.}, vol.~26, no.~6, pp. 1287--1299, Jun. 2015.

\bibitem{Elhamifar2009}
E.~Elhamifar and R.~Vidal, ``Sparse subspace clustering,'' in \emph{Proc. of
  22th IEEE Conf. Comput. Vis. and Pattern Recognit.}, Miami, FL, Jun. 2009,
  pp. 2790--2797.

\bibitem{Elhamifar2012}
------, ``Sparse subspace clustering: Algorithm, theory, and applications,''
  \emph{IEEE Trans. Pattern Anal. Mach. Intell.}, vol.~35, no.~11, pp.
  2765--2781, 2013.

\bibitem{Cheng2010}
B.~Cheng, J.~Yang, S.~Yan, Y.~Fu, and T.~Huang, ``Learning with
  $\ell^{1}$-graph for image analysis,'' \emph{IEEE Trans. on Image Process.},
  vol.~19, no.~4, pp. 858--866, 2010.

\bibitem{Xu2012}
D.~Xu, Y.~Huang, Z.~Zeng, and X.~Xu, ``Human gait recognition using patch
  distribution feature and locality-constrained group sparse representation,''
  \emph{IEEE Trans. on Image Process.}, vol.~21, no.~1, pp. 316--326, Jan.
  2012.

\bibitem{Jing2013}
L.~Jing, M.~Ng, and T.~Zeng, ``Dictionary learning-based subspace structure
  identification in spectral clustering,'' \emph{IEEE Trans. Neural. Netw.
  Learn. Syst.}, vol.~24, no.~8, pp. 1188--1199, Aug. 2013.

\bibitem{Gao2013PAMI}
S.~Gao, I.-H. Tsang, and L.-T. Chia, ``Laplacian sparse coding, hypergraph
  laplacian sparse coding, and applications,'' \emph{IEEE Trans. Pattern Anal.
  Mach. Intell.}, vol.~35, no.~1, pp. 92--104, Jan. 2013.

\bibitem{Liu2010}
G.~Liu, Z.~Lin, and Y.~Yu, ``Robust subspace segmentation by low-rank
  representation,'' in \emph{Proc. of 27th Int. Conf. Mach. Learn.}, Haifa,
  Israel, Jun. 2010, pp. 663--670.

\bibitem{Liu2012}
G.~Liu, Z.~Lin, S.~Yan, J.~Sun, Y.~Yu, and Y.~Ma, ``Robust recovery of subspace
  structures by low-rank representation,'' \emph{IEEE Trans. Pattern Anal.
  Mach. Intell.}, vol.~35, no.~1, pp. 171--184, 2013.

\bibitem{Favaro2011}
P.~Favaro, R.~Vidal, and A.~Ravichandran, ``A closed form solution to robust
  subspace estimation and clustering,'' in \emph{Proc. of 24th IEEE Conf.
  Comput. Vis. and Pattern Recognit.}, Colorado Springs, CO, Jun. 2011, pp.
  1801--1807.

\bibitem{Xiao2014}
S.~Xiao, M.~Tan, and D.~Xu, ``Weighted block-sparse low rank representation for
  face clustering in videos,'' in \emph{Proc. of 13th Eur. Conf. Comput. Vis.},
  2014, pp. 123--138.

\bibitem{Liu2011}
G.~C. Liu and S.~C. Yan, ``Latent low-rank representation for subspace
  segmentation and feature extraction,'' in \emph{Proc. of 13th IEEE Conf.
  Comput. Vis.}, Barcelona, Spain, Jun. 2011, pp. 1615--1622.

\bibitem{Lu2012}
C.-Y. Lu, H.~Min, Z.-Q. Zhao, L.~Zhu, D.-S. Huang, and S.~Yan, ``Robust and
  efficient subspace segmentation via least squares regression,'' in
  \emph{Proc. of 12th Eur. Conf. Comput. Vis.}, Florence, Italy, Oct. 2012, pp.
  347--360.

\bibitem{peng2015robust}
X.~Peng, Z.~Yi, and H.~Tang, ``Robust subspace clustering via thresholding
  ridge regression,'' in \emph{Proc. of 29th AAAI Conf. Artif. Intell.},
  Austin, TX, Jan. 2015, pp. 3827--3833.

\bibitem{Liu2012FRR}
R.~Liu, Z.~Lin, F.~D. la~Torre, and Z.~Su, ``Fixed-rank representation for
  unsupervised visual learning,'' in \emph{Proc. of 25th IEEE Conf. Comput.
  Vis. and Pattern Recognit.}, Providence, RI, Jun. 2012, pp. 598--605.

\bibitem{Peng2013SSSC}
X.~Peng, L.~Zhang, and Z.~Yi, ``Scalable sparse subspace clustering,'' in
  \emph{Proc. of 26th IEEE Conf. Comput. Vis. and Pattern Recognit.}, Portland,
  OR, Jun. 2013, pp. 430--437.

\bibitem{Wright2009}
J.~Wright, A.~Y. Yang, A.~Ganesh, S.~S. Sastry, and Y.~Ma, ``Robust face
  recognition via sparse representation,'' \emph{IEEE Trans. Pattern Anal.
  Mach. Intell.}, vol.~31, no.~2, pp. 210--227, 2009.

\bibitem{Yang2010}
A.~Yang, A.~Ganesh, S.~Sastry, and Y.~Ma, ``Fast l1-minimization algorithms and
  an application in robust face recognition: A review,'' EECS Department,
  University of California, Berkeley, Tech. Rep. UCB/EECS-2010-13, Feb. 2010.

\bibitem{Xiao_2015_CVPR}
S.~Xiao, W.~Li, D.~Xu, and D.~Tao, ``Fa{LRR}: A fast low rank representation
  solver,'' in \emph{Proc. of 28th IEEE Conf. Comput. Vis. and Pattern
  Recognit.}, Boston, MA, Jun. 2015, pp. 4612--4620.

\bibitem{Liu2012AS}
G.~Liu and S.~Yan, ``Active subspace: Toward scalable low-rank learning,''
  \emph{Neural Comput.}, vol.~24, no.~12, pp. 3371--3394, 2012.

\bibitem{Zhang2013}
X.~Zhang, F.~Sun, G.~Liu, and Y.~Ma, ``Fast low-rank subspace segmentation,''
  \emph{IEEE Trans. Knowl. Data Eng.}, vol.~26, no.~5, pp. 1293--1297, May
  2014.

\bibitem{Fowlkes2004}
C.~Fowlkes, S.~Belongie, F.~Chung, and J.~Malik, ``Spectral grouping using the
  nystrom method,'' \emph{IEEE Trans. Pattern Anal. Mach. Intell.}, vol.~26,
  no.~2, pp. 214--225, 2004.

\bibitem{Chen2011a}
W.-Y. Chen, Y.~Song, H.~Bai, C.-J. Lin, and E.~Y. Chang, ``Parallel spectral
  clustering in distributed systems,'' \emph{IEEE Trans. Pattern Anal. Mach.
  Intell.}, vol.~33, no.~3, pp. 568--586, 2011.

\bibitem{Yan2009}
D.~Yan, L.~Huang, and M.~I. Jordan, ``Fast approximate spectral clustering,''
  in \emph{Proc. of 15th {ACM SIGKDD} Int. Conf. Knowl. Dis. and Data Min.},
  Paris, France, Jun. 2009, pp. 907--916.

\bibitem{Chen2011}
X.~Chen and D.~Cai, ``Large scale spectral clustering with landmark-based
  representation,'' in \emph{Proc. of 25th AAAI Conf. Artif. Intell.}, San
  Francisco, CA, Aug. 2011, pp. 313--318.

\bibitem{Wang2011}
L.~Wang, C.~Leckie, R.~Kotagiri, and J.~Bezdek, ``Approximate pairwise
  clustering for large data sets via sampling plus extension,'' \emph{Pattern
  Recogn.}, vol.~44, no.~2, pp. 222--235, 2011.

\bibitem{Nie2011}
F.~Nie, Z.~Zeng, T.~I. W., D.~Xu, and C.~Zhang, ``Spectral embedded clustering:
  A framework for in-sample and out-of-sample spectral clustering,'' \emph{IEEE
  Trans. Neural. Netw.}, vol.~22, no.~11, pp. 1796--1808, 2011.

\bibitem{Belabbas2009}
M.-A. Belabbas and P.~J. Wolfe, ``Spectral methods in machine learning and new
  strategies for very large datasets,'' \emph{Proc. of Natl. Acad. Sci.}, vol.
  106, no.~2, pp. 369--374, 2009.

\bibitem{Tasdemir2012}
K.~Tasdemir, ``Vector quantization based approximate spectral clustering of
  large datasets,'' \emph{Pattern Recogn.}, vol.~45, no.~8, pp. 3034--3044,
  2012.

\bibitem{Halko2011}
N.~Halko, P.~Martinsson, and J.~Tropp, ``Finding structure with randomness:
  Probabilistic algorithms for constructing approximate matrix
  decompositions,'' \emph{{SIAM} Review}, vol.~53, no.~2, pp. 217--288, 2011.

\bibitem{Zhang2011}
L.~Zhang, M.~Yang, and X.~Feng, ``Sparse representation or collaborative
  representation: Which helps face recognition?'' in \emph{Proc. of IEEE Int.
  Conf. on Comput. Vis.}, Barcelona, Spain, Nov. 2011, pp. 471--478.

\bibitem{Naseem2010}
I.~Naseem, R.~Togneri, and M.~Bennamoun, ``Linear regression for face
  recognition,'' \emph{IEEE Trans. Pattern Anal. Mach. Intell.}, vol.~32,
  no.~11, pp. 2106--2112, Nov. 2010.

\bibitem{Smith2010}
M.~Smith, N.~Milic-Frayling, B.~Shneiderman, E.~Mendes~Rodrigues, J.~Leskovec,
  and C.~Dunne, ``Nodexl: a free and open network overview, discovery and
  exploration add-in for excel 2007/2010,'' \emph{Social Media Research
  Foundation}, 2010.

\bibitem{Luxburg2004}
U.~V. Luxburg, O.~Bousquet, and M.~Belkin, ``Limits of spectral clustering,''
  in \emph{Proc. of 17th Adv. in Neural Inf. Process. Syst.}, Hyatt Regency,
  Canada, Dec. 2004, pp. 857--864.

\bibitem{Gao2013TIP}
S.~Gao, I.~W.-H. Tsang, and L.-T. Chia, ``Sparse representation with kernels,''
  \emph{IEEE Trans. on Image Process.}, vol.~22, no.~2, pp. 423--434, 2013.

\bibitem{Wang2013}
Z.~Wang, J.~Yang, N.~Nasrabadi, and T.~Huang, ``A max-margin perspective on
  sparse representation-based classification,'' in \emph{Proc. of IEEE Conf.
  Comput. Vis.}, Sydney, Australia, Dec. 2013, pp. 1217--1224.

\bibitem{Hamm2008}
J.~Hamm and D.~D. Lee, ``Grassmann discriminant analysis: a unifying view on
  subspace-based learning,'' in \emph{Proc. of 25th Int. Conf. Mach. Learn.},
  Helsinki, Finland, Jul. 2008, pp. 376--383.

\bibitem{Fukunaga1990}
K.~Fukunaga, \emph{Introduction to Statistical Pattern Recogn. (2nd
  Ed.)}.\hskip 1em plus 0.5em minus 0.4em\relax San Diego, CA: Academic Press
  Professional, Inc., 1990.

\bibitem{Chitta2011}
R.~Chitta, R.~Jin, T.~Havens, and A.~Jain, ``Approximate kernel k-means:
  solution to large scale kernel clustering,'' in \emph{Proc. of 17th {ACM
  SIGKDD} Int. Conf. Knowl. Dis. and Data Min.}, San Diego, CA, Aug. 2011, pp.
  895--903.

\bibitem{Georghiades2001}
A.~S. Georghiades, P.~N. Belhumeur, and D.~J. Kriegman, ``From few to many:
  Illumination cone models for face recognition under variable lighting and
  pose,'' \emph{IEEE Trans. Pattern Anal. Mach. Intell.}, vol.~23, no.~6, pp.
  643--660, 2001.

\bibitem{Osborne2000}
M.~R. Osborne, B.~Presnell, and B.~A. Turlach, ``A new approach to variable
  selection in least squares problems,'' \emph{{IMA} Journal of Numerical
  Analysis}, vol.~20, no.~3, pp. 389--403, 2000.

\bibitem{Martinez1998}
A.~Martinez, ``The {AR} face database,'' \emph{CVC Technical Report}, vol.~24,
  1998.

\bibitem{Gary2007}
G.~B. Huang, M.~Ramesh, T.~Berg, and E.~Learned-Miller, ``Labeled faces in the
  wild: A database for studying face recognition in unconstrained
  environments,'' University of Massachusetts, Amherst, Tech. Rep. 07--49, Oct.
  2007.

\bibitem{Gross2010}
R.~Gross, I.~Matthews, J.~Cohn, T.~Kanade, and S.~Baker, ``Multi-{PIE},''
  \emph{Image Vision Comput.}, vol.~28, no.~5, pp. 807--813, 2010.

\bibitem{Yang2012-Relaxed}
M.~Yang, L.~Zhang, D.~Zhang, and S.~Wang, ``Relaxed collaborative
  representation for pattern classification,'' in \emph{Proc. of 25th IEEE
  Conf. Comput. Vis. and Pattern Recognit.}, Providence, RI, Jun. 2012, pp.
  2224--2231.

\bibitem{Cai2005}
D.~Cai, X.~F. He, and J.~W. Han, ``Document clustering using locality
  preserving indexing,'' \emph{IEEE Trans. Knowl. Data En.}, vol.~17, no.~12,
  pp. 1624--1637, 2005.

\bibitem{Alimoglu1996}
F.~Alimoglu and E.~Alpaydin, ``Combining multiple representations and
  classifiers for pen-based handwritten digit recognition,'' in \emph{Proc. of
  4th Int. Conf. Doc. Anal. and Recognit.}, ULM, Germany, Aug. 1997, pp.
  637--640.

\bibitem{Blackard1999}
J.~Blackard and D.~Dean, ``Comparative accuracies of artificial neural networks
  and discriminant analysis in predicting forest cover types from cartographic
  variables,'' \emph{Comput. Electron. Agric.}, vol.~24, no.~3, pp. 131--151,
  1999.

\bibitem{Taigman2009}
Y.~Taigman, L.~Wolf, and T.~Hassner, ``Multiple one-shots for utilizing class
  label information.'' in \emph{Proc. of 20th Brit. Mach. Vis. Conf.}, London,
  England, Sep. 2009, pp. 1--12.

\bibitem{Filippone2008}
M.~Filippone, F.~Camastra, F.~Masulli, and S.~Rovetta, ``A survey of kernel and
  spectral methods for clustering,'' \emph{Pattern Recogn.}, vol.~41, no.~1,
  pp. 176--190, 2008.

\bibitem{Cai2011kmeans}
D.~Cai, ``Litekmeans: the fastest matlab implementation of kmeans,''
  \emph{Available at:
  \url{http://www.zjucadcg.cn/dengcai/Data/Clustering.html}}, 2011.

\bibitem{Zhao2001}
Y.~Zhao and G.~Karypis, ``Empirical and theoretical comparisons of selected
  criterion functions for document clustering,'' \emph{Mach. Learn.}, vol.~55,
  no.~3, pp. 311--331, 2004.

\bibitem{negahban2011estimation}
S.~Negahban and M.~J. Wainwright, ``Estimation of (near) low-rank matrices with
  noise and high-dimensional scaling,'' \emph{Ann. Stat.}, vol.~39, no.~2, pp.
  1069--1097, 2011.

\bibitem{Kvalseth1987}
T.~O. Kvalseth, ``Entropy and correlation: Some comments,'' \emph{IEEE Trans.
  Syst. Man Cybern.}, vol.~17, no.~3, pp. 517--519, 1987.

\end{thebibliography}



%

%

\begin{IEEEbiography}[{\includegraphics[width=1in,height=1.25in,clip,keepaspectratio]{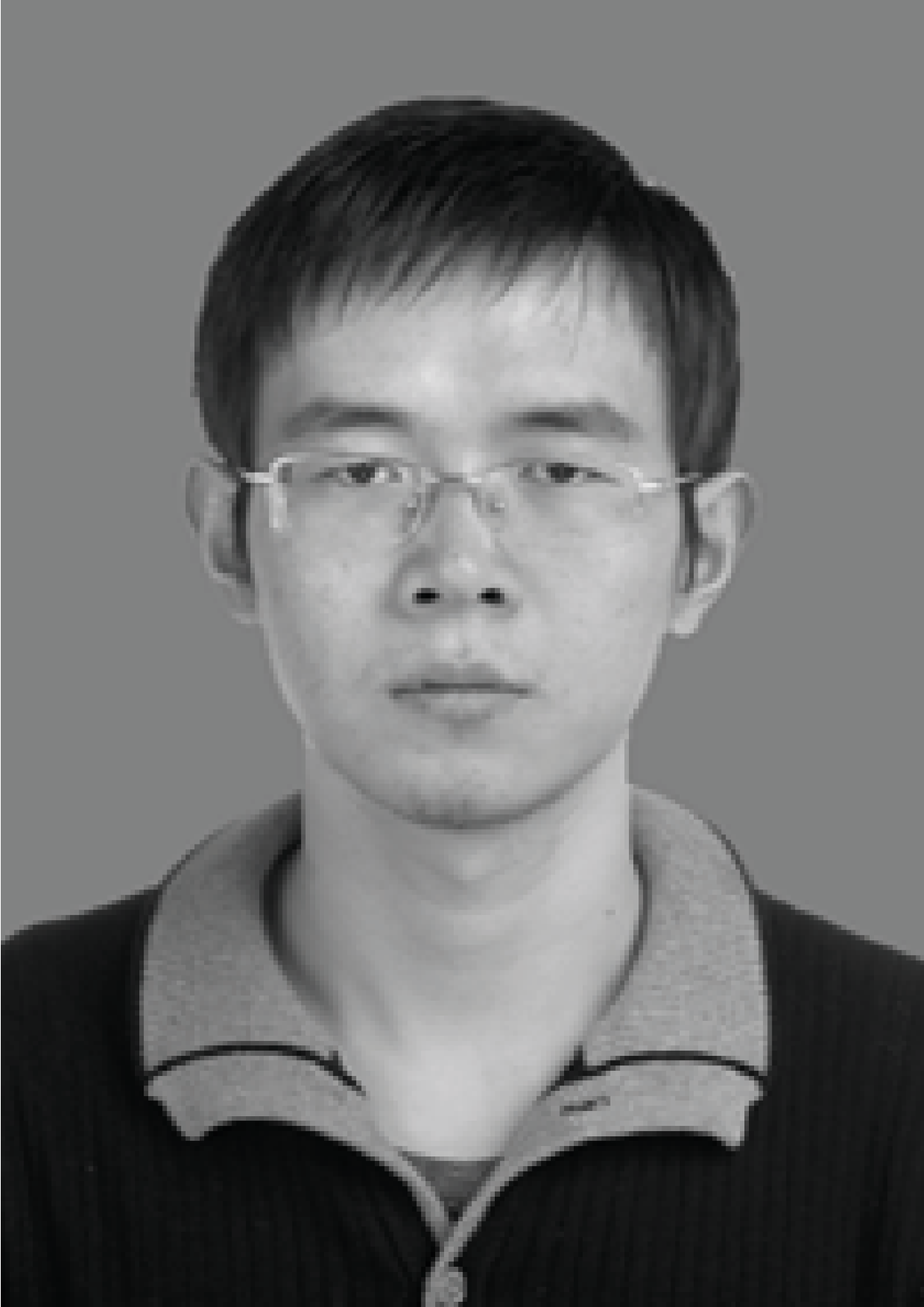}}]{Xi Peng}
is a research scientist at Institute for Infocomm., Research Agency for Science, Technology and Research (A*STAR) Singapore. He received the BEng degree in Electronic Engineering and MEng degree in Computer Science from Chongqing University of Posts and Telecommunications, and the Ph.D. degree from Sichuan University, China, respectively. His current research interests include computer
vision, image processing, and pattern recognition. 

 Dr. Peng is the recipient of China National Graduate Scholarship in 2013, CSC-IBM Scholarship for Outstanding Chinese Students in 2012, and Excellent Student Paper of IEEE CHENGDU Section in 2010. He has served as a PC member for 10 international conferences such as IJCNN 2014-2016 and a reviewer for over 10 international journals such as IEEE TNNLS, TIP, TKDE, TIFS, TGRS, TCYB.
\end{IEEEbiography}

\begin{IEEEbiography}[{\includegraphics[width=1in,height=1.25in,clip,keepaspectratio]{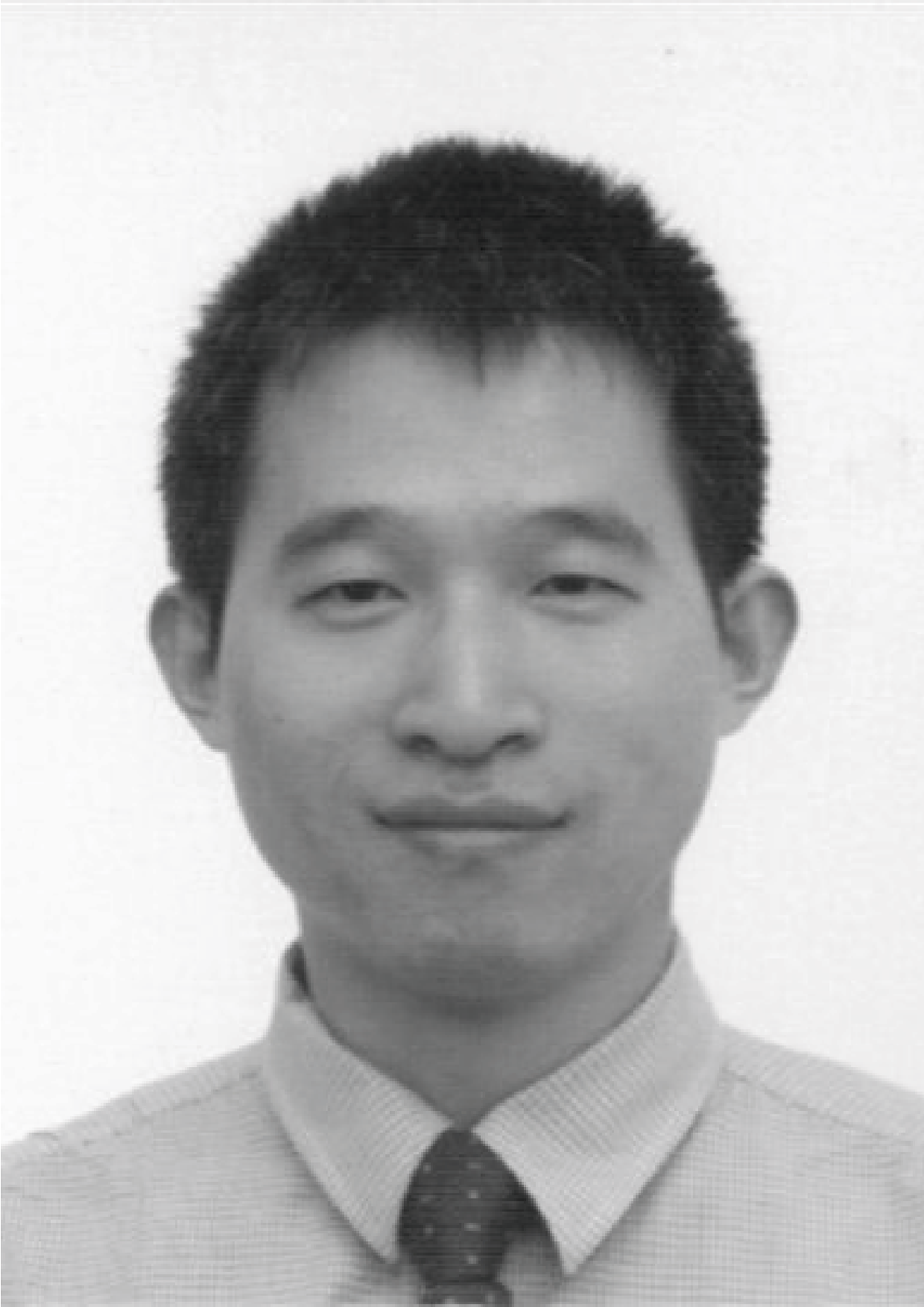}}]{Huajin Tang}
(M'01) received the B.Eng. degree from Zhejiang University, Hangzhou, China, in 1998, the M.Eng. degree from Shanghai Jiao Tong University, Shanghai, China, in 2001, and the Ph.D. degree in electrical and computer engineering from the National University of Singapore, Singapore, in 2005. He was a System Engineer with STMicroelectronics, Singapore, from 2004 to 2006, and then a Post-Doctoral Fellow with the Queensland Brain Institute, University of Queensland, Brisbane, QLD, Australia, from 2006 to 2008. He is currently a Research Scientist leading the Cognitive Computing Group with the Institute for Infocomm Research, Agency for Science, Technology and Research, Singapore. He has authored one monograph (Springer-Verlag, 2007) and over 30 international journal papers. His current research interests include neural computation, neuromorphic cognitive systems, neurocognitive robots, and machine learning. 

Dr. Tang serves as an Associate Editor of the \textit{IEEE TRANSACTIONS ON NEURAL NETWORKS AND LEARNING SYSTEMS} and an Editorial Board Member of Frontiers in Robotics and AI.
\end{IEEEbiography}


\begin{IEEEbiography}[{\includegraphics[width=1in,height=1.25in,clip,keepaspectratio]{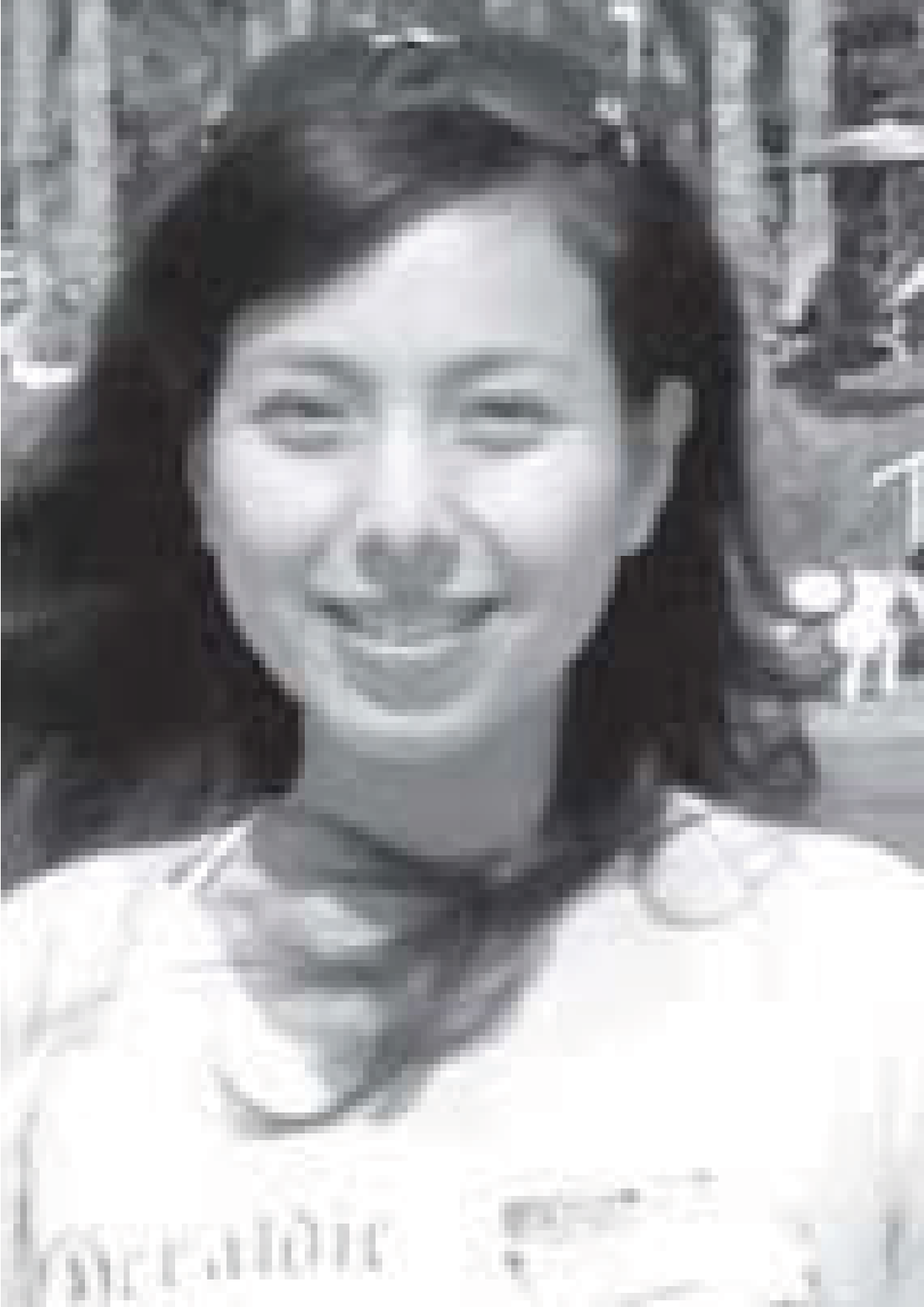}}]{Lei Zhang} (M'10)
received the B.S. and Masters degrees in mathematics and the Ph.D. degree in computer science from the University of Electronic Science and Technology of China, Chengdu, China, in 2002, 2005, and 2008, respectively. She was a Post-Doctoral Research Fellow in the Department of Computer Science and Engineering, Chinese University of Hong Kong, Shatin, Hong Kong, from 2008 to 2009. Currently, she is a Professor at Sichuan University, Chengdu. Her current research interests include theory and applications of neural networks based on neocortex computing and big data analysis methods by infinity deep neural networks.
\end{IEEEbiography}

\begin{IEEEbiography}[{\includegraphics[width=1in,height=1.25in,clip,keepaspectratio]{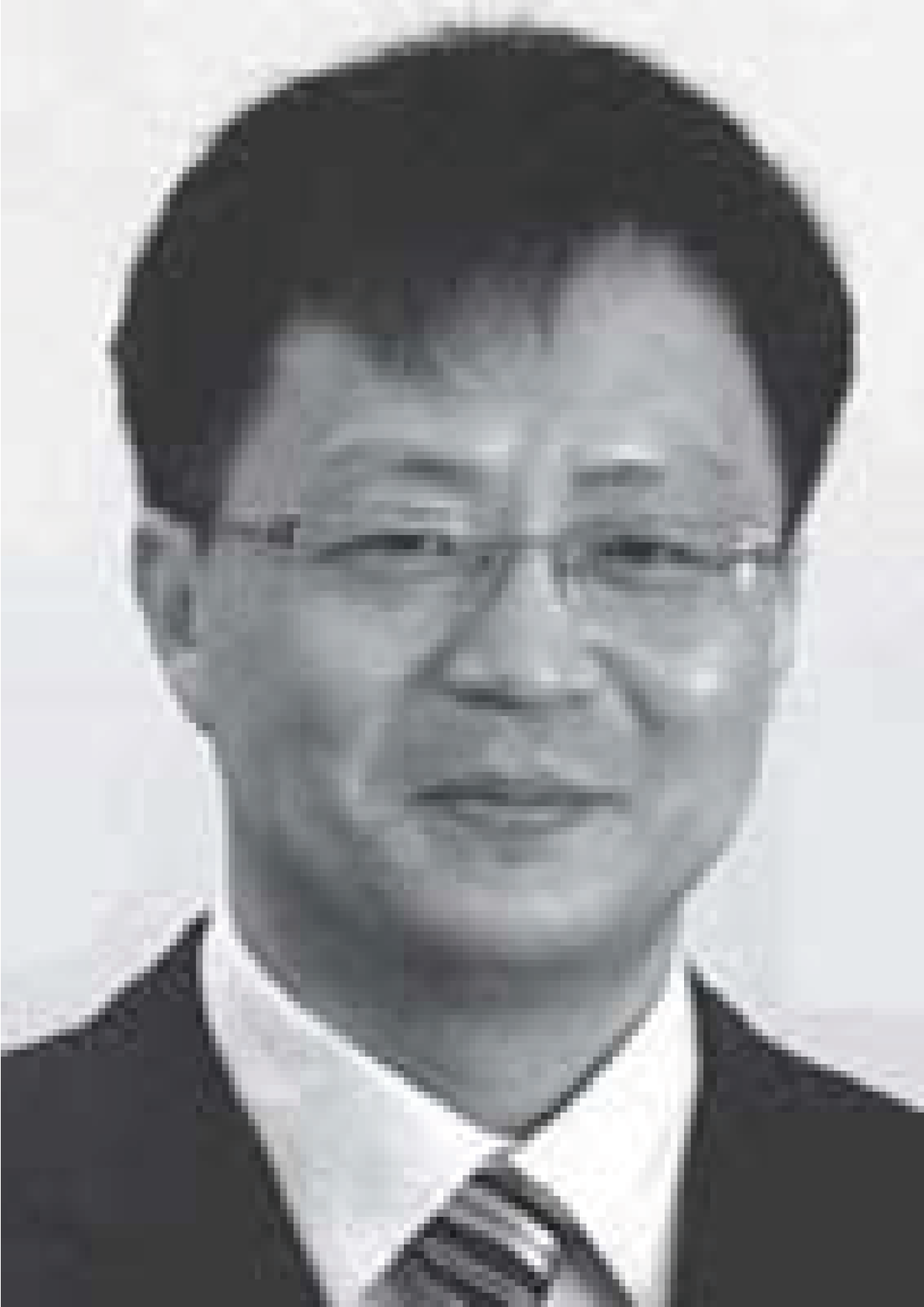}}]{Zhang Yi} (SM'10)
received the Ph.D. degree in mathematics from the Institute of Mathematics, The Chinese Academy of Science, Beijing, China, in 1994. Currently, he is a Professor at the College of Computer Science, Sichuan University, Chengdu, China. He is the co-author of three books: Convergence Analysis of Recurrent Neural Networks (Kluwer Academic Publisher, 2004), Neural Networks: Computational Models and Applications (Springer, 2007), and Subspace Learning of Neural Networks (CRC Press, 2010). He is the Chair of IEEE Chengdu Section (2015 ~). He was an Associate Editor of IEEE Transactions on Neural Networks and Learning Systems (2009 ~ 2012), and an Associate Editor of IEEE Transactions on Cybernetics (2014 ~). His current research interests include Neural Networks and Big Data. He is the founding director of Machine Intelligence Laboratory. He is also the founder of IEEE Computational Intelligence Society, Chengdu Chapter.
\end{IEEEbiography}

\begin{IEEEbiography}[{\includegraphics[width=1in,height=1.25in,clip,keepaspectratio]{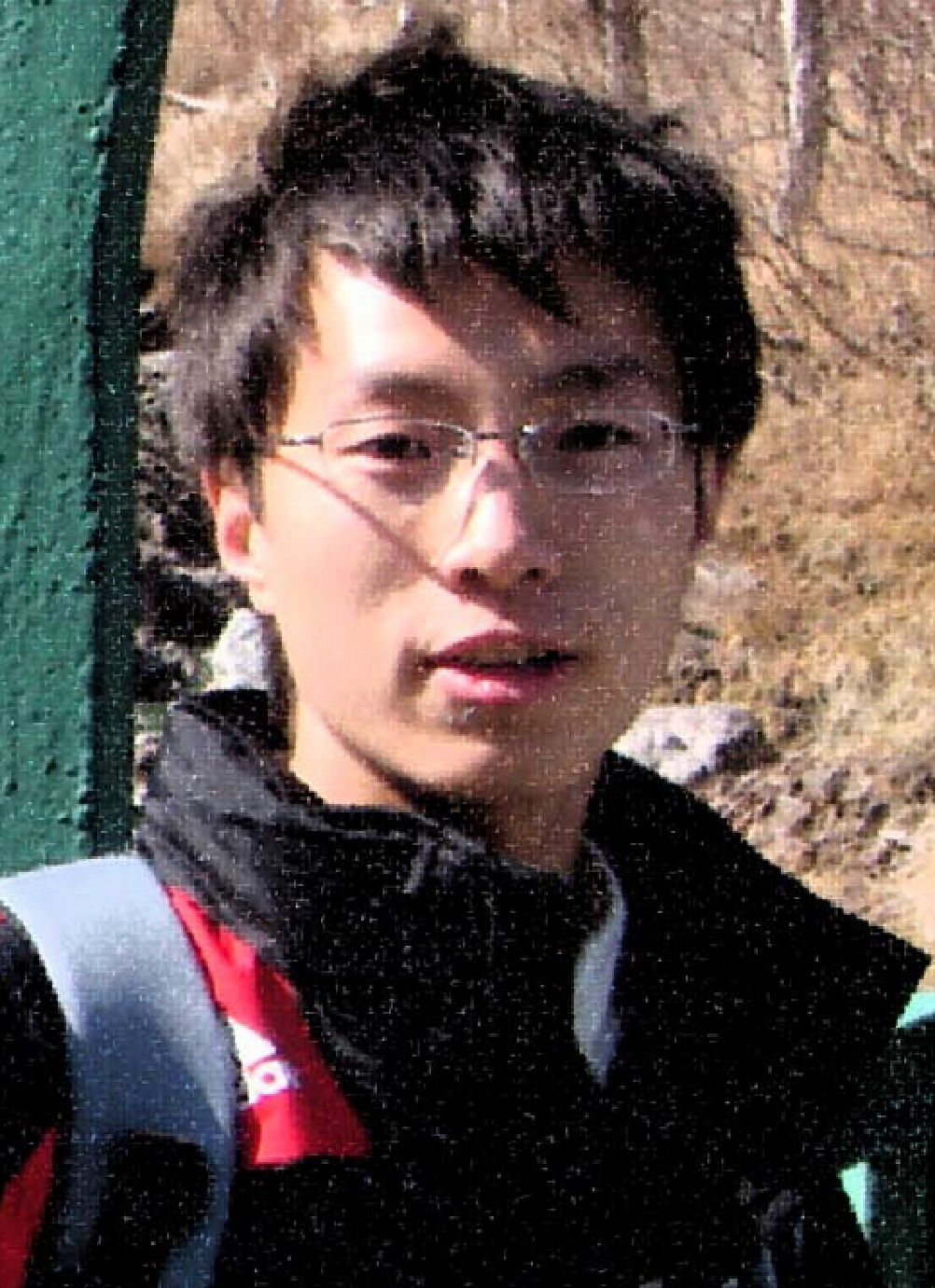}}]{Shijie Xiao}
	received the B.E. degree from the Harbin Institute of Technology, Harbin, China, in 2011. He is currently pursuing the Ph.D. degree with the School of Computer Engineering, Nanyang Technological University, Singapore.
\\
$~~$
His current research interests include machine learning and computer vision.
\end{IEEEbiography}




\end{document}